\def\skipintrofigure{}

\documentclass{article}
\usepackage{iclr2027_conference,times}

\iclrfinalcopy

\usepackage{amsmath,amsfonts,bm}

\def\eqref#1{equation~\ref{#1}}

\def\1{\bm{1}}

\def\vx{{\bm{x}}}

\DeclareMathAlphabet{\mathsfit}{\encodingdefault}{\sfdefault}{m}{sl}
\SetMathAlphabet{\mathsfit}{bold}{\encodingdefault}{\sfdefault}{bx}{n}

\usepackage[hidelinks]{hyperref}
\usepackage{url}
\usepackage{booktabs}
\usepackage{multirow}
\usepackage{graphicx}
\usepackage{amsmath}
\usepackage{amssymb}
\usepackage{xcolor}
\usepackage{colortbl}
\usepackage{tabularx}
\usepackage{tcolorbox}
\tcbuselibrary{breakable}
\usepackage{makecell}
\usepackage{pifont}
\usepackage{enumitem}

\newcommand{\nexp}{E}                    
\newcommand{\ktop}{K}                    
\newcommand{\kbar}{\bar{k}}              
\newcommand{\rbudget}{\rho}              
\newcommand{\fixedk}{\textsc{Fixed-K}}
\newcommand{\methodname}[1]{\textsc{#1}}
\newcommand{\naee}{\methodname{Naee}}
\newcommand{\diep}{\methodname{DiEP}}
\newcommand{\ban}{\methodname{Ban}}
\newcommand{\dynr}{\methodname{DynRoute}}

\definecolor{takeawaybg}{HTML}{F1F5FA}
\definecolor{takeawayrule}{HTML}{31597F}

\newtcolorbox{takeaway}[1][]{
  colback=takeawaybg, colframe=takeawayrule,
  boxrule=0pt, leftrule=2.4pt, sharp corners,
  left=8pt, right=8pt, top=5pt, bottom=5pt,
  breakable, #1}

\newcommand{\finding}[2]{\noindent\textcolor{takeawayrule}{\textbf{#1}}~\emph{#2}}

\newtcolorbox{questions}[1][]{
  colback=takeawaybg, colframe=takeawayrule,
  boxrule=0.9pt, sharp corners,
  left=10pt, right=10pt, top=3pt, bottom=3pt,
  breakable, #1}

\newlist{flushitems}{itemize}{1}
\setlist[flushitems]{label=\textbullet, leftmargin=*, labelsep=0.45em,
                     topsep=6pt, itemsep=7pt, parsep=0pt}

\definecolor{modelfill}{HTML}{E6E6E6}
\definecolor{unprunedfill}{HTML}{EFEFEF}   
\definecolor{groupfill}{HTML}{E4EBF4}
\definecolor{groupink}{HTML}{2B4F70}
\definecolor{consfill}{HTML}{E8F1EA}
\definecolor{aggrfill}{HTML}{FBEDE4}

\newcolumntype{L}{>{\raggedright\arraybackslash}X}

\newcommand{\yes}{\ding{51}}
\newcommand{\no}{\ding{55}}

\newcommand{\arxivnameline}{%
  Yuanteng Chen\textsuperscript{1,2,3,*},
  Qiwei Lai\textsuperscript{4,*},
  Chen Tianqi\textsuperscript{1,2,*},
  Peisong Wang\textsuperscript{1,2,\textdagger},
  Yuantian Shao\textsuperscript{1},\\
  Nanxin Zeng\textsuperscript{2},
  Zhilei Liu\textsuperscript{1,2},
  Chuangyi Li\textsuperscript{1,2},
  Jing Liu\textsuperscript{1,2,3},
  Jian Cheng\textsuperscript{1,2,3,\textdagger}}

\newcommand{\arxivaffilblock}{%
  \textsuperscript{1}\,Institute of Automation, Chinese Academy of Sciences\\
  \textsuperscript{2}\,School of Artificial Intelligence, University of Chinese Academy of Sciences\\
  \textsuperscript{3}\,Zhongguancun Academy
  \quad
  \textsuperscript{4}\,University of Science and Technology of China\\
  \textsuperscript{*}\,Equal contribution
  \quad
  \textsuperscript{\textdagger}\,Corresponding authors}

\makeatletter
\def\@maketitle{\vbox{\hsize\textwidth
\parskip=0pt
{\centering\LARGE\sc \@title\par}
\vskip 8pt
{\centering\normalsize\bfseries\arxivnameline\par}
\vskip 6pt
{\centering\normalsize\arxivaffilblock\par}
\vskip 8pt}}
\makeatother
\renewenvironment{abstract}{\vskip 2pt\centerline{\large\sc
Abstract}\vspace{0.5ex}\noindent\ignorespaces}{\par\vskip 2pt}

\title{You Only Need 2/3 of the Chosen Experts:\\ An Empirical Study of Dynamic Expert\\ Pruning in Fine-Grained MoE LLMs}

\author{Yuanteng Chen}

\fancypagestyle{p1teaser}{%
  \fancyhf{}%
  \fancyhead[L]{\small\scshape Preprint}%
  \renewcommand{\headrulewidth}{0.4pt}%
}

\begin{document}

\thispagestyle{p1teaser}
\maketitle
\enlargethispage*{88pt}

\begin{abstract}
Fine-grained mixture-of-experts (MoE) architectures have become a mainstream design for open-weight
LLMs, with hundreds of experts and increasingly many selected per token. This shift makes dynamic
expert pruning an attractive route to cheaper inference. Yet existing evidence comes largely from
coarser architectures and likelihood-scored multiple-choice benchmarks, leaving three central questions
insufficiently understood in the fine-grained regime: how redundant per-token expert selection is, how
effectively existing pruning methods exploit that redundancy, and what governs a model's sensitivity to
pruning. We fill this gap with a systematic empirical study of twelve fine-grained MoE checkpoints
spanning nine architecture families, with a core suite of eleven benchmarks covering knowledge QA,
mathematics, code generation, and general reasoning. We find that expert selection is far more redundant
than the field's operating points assume: uniformly retaining approximately two thirds of the selected
experts preserves $98.8\%$ of unpruned performance on average, requiring only a one-integer change and
delivering $1.2$ to $1.7\times$ measured speedup across two serving backends.
This simple baseline leaves little room for dynamic allocation at conservative budgets: even the best
published rules differ from it by less than one percentage point at matched average expert counts. Their
value emerges under aggressive pruning, where the best rules recover up to $3.0$ points over uniform
truncation, with gains concentrated in the generative tasks that suffer the sharpest degradation. This
sensitivity to aggressive pruning also depends on the model: larger and thinking models are more
resilient, whereas multimodal models are more vulnerable. Together, these findings reveal how much
expert computation
fine-grained MoE models can dispense with, and establish when dynamic allocation earns its complexity,
informing both practical deployment and future pruning methods.
Code is available at~\href{https://github.com/MingZwhy/An-Empirical-Study-on-Expert-Pruning}{\raisebox{-0.18ex}{\includegraphics[height=1.05em]{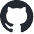}}\,GitHub}.
\end{abstract}

\par\nointerlineskip
\vspace{2pt}
\noindent\includegraphics[width=\linewidth]{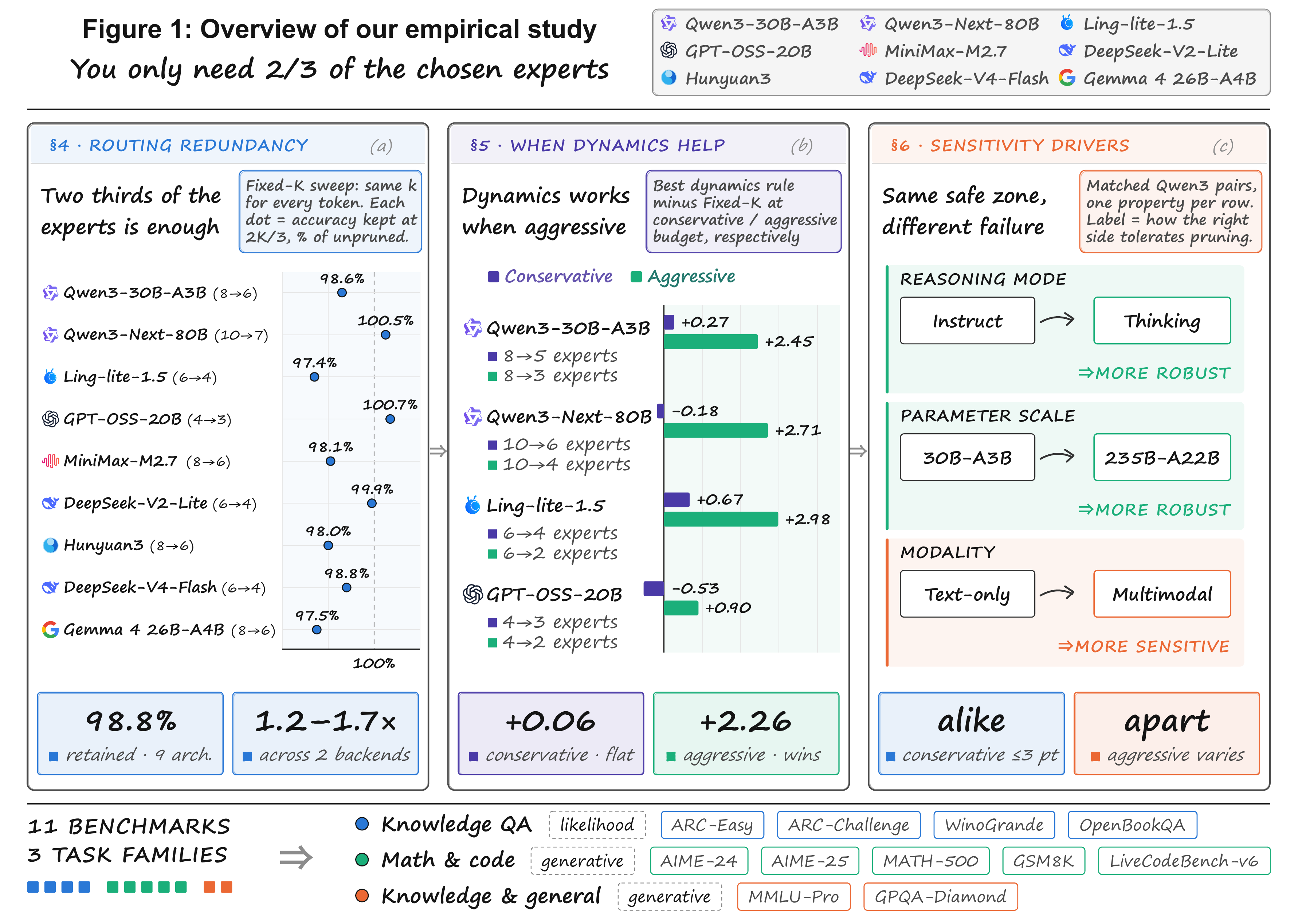}%
\refstepcounter{figure}\label{fig:teaser}
\clearpage

\section{Introduction}
\label{sec:intro}

Scaling a language model has long meant paying for every parameter on every token. Mixture-of-experts
(MoE) layers break that coupling, allowing model capacity and per-token computation to scale
independently. Fine-grained MoE has since become a mainstream design for open-weight LLMs: expert pools
have grown into the hundreds, and the number selected per token has grown with them. Mixtral selects
$2$ of $8$ experts; Qwen3-30B-A3B selects $8$ of $128$, and Qwen3-Next-80B-A3B selects $10$ of $512$.
This evolution makes routing a richer allocation problem, but also raises a basic question: how much of
the computation selected by these routers is actually necessary?

\ifdefined\skipintrofigure\else
\begin{figure}[t]
\centering
\includegraphics[width=\linewidth]{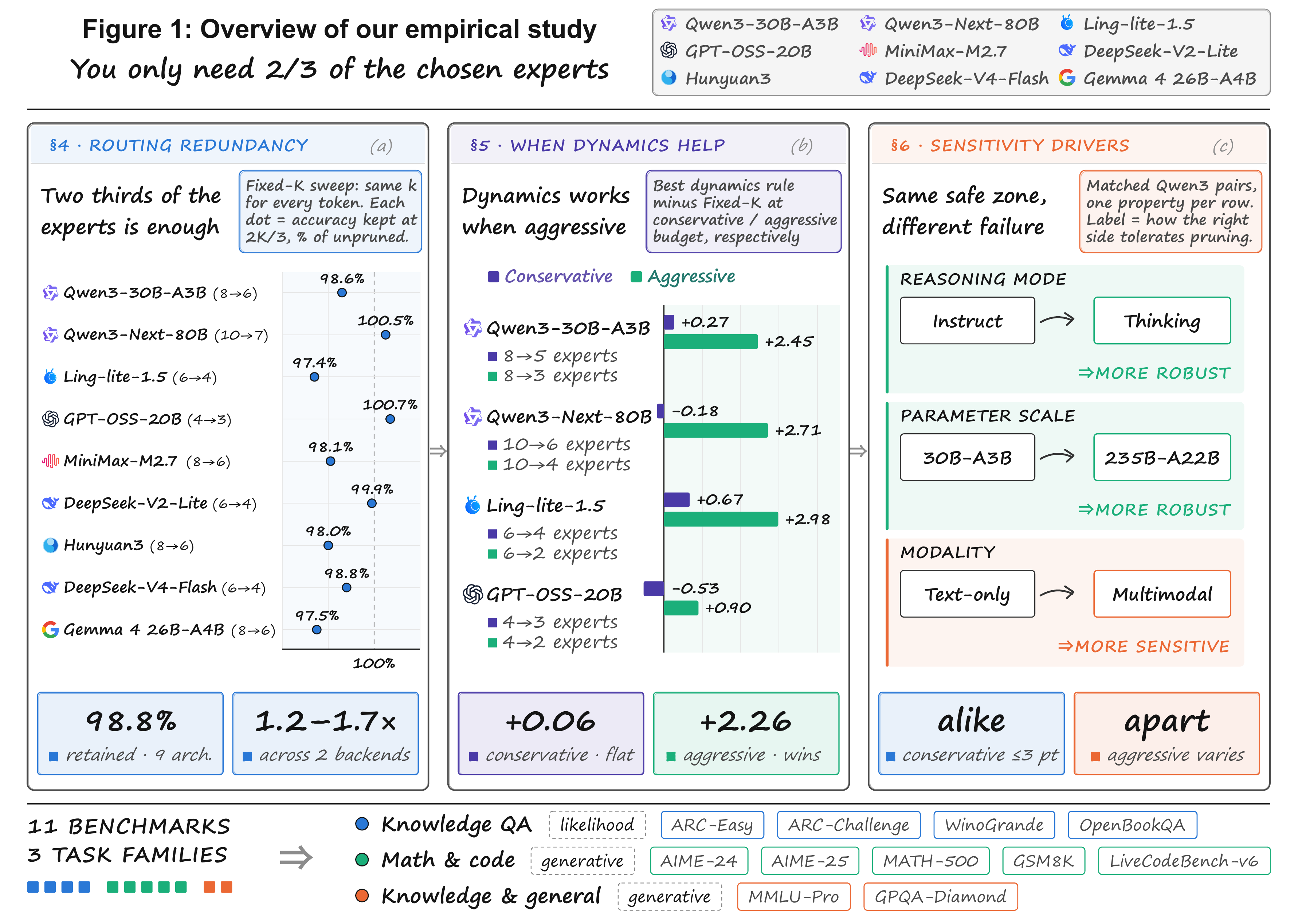}
\caption{Overview. \textbf{(a)} Uniformly retaining approximately two thirds of the selected experts
preserves $98.8\%$ of unpruned performance on average. \textbf{(b)} The best dynamic rules add little
over uniform truncation at conservative budgets but recover significantly under aggressive pruning.
\textbf{(c)} Paired comparisons show that thinking and larger models are more robust, whereas the
multimodal model is more sensitive.}
\vspace{-8pt}
\label{fig:teaser}
\end{figure}
\fi

A top-$\ktop$ router assigns the same expert count to every token, whether its scores are sharply peaked
or spread across the selected experts. Dynamic expert pruning makes that count adaptive, using the
router's scores to decide how many experts to execute for each token and layer. Fine-grained
architectures appear especially suited to this approach: a longer ranked list offers more freedom to
prune, and every skipped expert removes an entire feed-forward evaluation. The promise is
straightforward: spend expert computation where it matters most. Realising that promise, however,
requires knowing how much of the original budget was needed in the first place.

Existing studies have not established that picture for fine-grained MoE. Most pruning rules were
developed on coarser architectures and evaluated primarily on likelihood-scored multiple-choice
benchmarks. Ranking candidate answers reveals little about whether the same pruning budget preserves a
long mathematical derivation or a working program. Methods are also reported in isolation, on different
models and at different budgets, so their results do not separate the redundancy already present in the
model from the additional benefit of adaptive allocation. What the field lacks is a systematic account
of both: how much expert selection can be removed, and when a dynamic rule is needed to remove it. We
address this gap through three questions:

\begin{questions}
\begin{enumerate}[label=\textbf{Q\arabic*.},leftmargin=2.6em,itemsep=1.5pt,topsep=1pt,parsep=0pt]
\item How much redundancy does per-token expert selection carry in fine-grained MoE?
\item When does dynamic allocation improve on simply lowering the expert count?
\item What governs a model's sensitivity to expert pruning?
\end{enumerate}
\end{questions}

We study twelve checkpoints spanning nine architecture families. Nine models establish the redundancy of
expert selection through uniform budget sweeps; four of them also support comparisons of published
dynamic rules, and three additional checkpoints probe reasoning mode, parameter scale and modality. Our
core evaluation combines four likelihood-scored QA benchmarks with seven generative benchmarks covering
mathematics, code and general reasoning. This breadth lets us examine both the cost of pruning and the
extent to which that cost depends on what a model is asked to do.

The study yields three central findings, summarised in Figure~\ref{fig:teaser}:

\begin{flushitems}
\item \textbf{A systematic characterisation of routing redundancy.} We show that fine-grained expert
selection is far more redundant than the field's operating points assume. Uniformly retaining approximately two thirds of the selected experts preserves $98.8\%$ of unpruned
performance on average across the nine models and three task families. This requires only lowering a
single integer: no calibration, no token-specific allocation, and no change to the router's ranking. Much
of the apparent opportunity for dynamic pruning is therefore already captured by the simplest
baseline.

\item \textbf{A reassessment of dynamic pruning across budgets.} We establish when dynamic allocation
earns its complexity beyond simply lowering the expert count. Against this uniform baseline at
matched average expert counts, even the best published rules add little under conservative pruning,
with margins between $-0.53$ and $+0.67$ percentage points. Their advantage emerges once pruning becomes
aggressive, where the best rules recover up to $3.0$ points and the winning rule changes across models
and budgets. The central question for a pruning method is therefore where it improves the
quality--compute trade-off, not whether it can match uniform truncation.

\item \textbf{Sensitivity analysis and actionable guidance.} We identify how task and model properties
shape pruning sensitivity, and translate these findings into guidance for deployment and future method
design. Generative tasks deteriorate more sharply than likelihood-scored QA and account for most of the gains
from dynamic allocation. Extending this analysis to model properties, we find that larger and thinking
models withstand aggressive pruning better, whereas multimodal models are more sensitive; these
differences are small at conservative budgets. Together, these results show why choosing a pruning
budget requires understanding both the capabilities being evaluated and the model being pruned.
\end{flushitems}

These findings establish a practical starting point for fine-grained MoE inference: capture the abundant
redundancy by lowering $\ktop$ uniformly, then use dynamic allocation to push further. For future
methods, the opportunity lies in improving that trade-off where uniform truncation begins to fail.

\section{Related work}
\label{sec:related}

\subsection{Fine-grained MoE}
\label{sec:related:moe}

MoE architectures \citep{shazeer2017outrageously, lepikhin2021gshard, fedus2022switch} have evolved
from routing among a few wide experts to combining many narrower ones.
Mixtral activates $2$ of $8$ experts per token, with each expert matching the width of a dense
feed-forward network \citep{jiang2024mixtral}. DeepSeekMoE introduced finer expert segmentation
together with shared experts, expanding the combinations available to each token
\citep{dai2024deepseekmoe}. This design has become mainstream in open-weight LLMs: Qwen3-30B-A3B
activates $8$ of $128$ routed experts \citep{qwen3} and Qwen3-Next-80B-A3B $10$ of $512$
\citep{qwen3next}, while GLM-5
and MiniMax-M2 each activate $8$ of $256$ \citep{glm5, minimaxm2}. Kimi K3 extends this trend to $16$
of $896$ routed experts \citep{kimik3}.

This evolution changes the scope of expert pruning. On a top-$2$ router, removing one expert eliminates
half of the selected computation; a longer selection offers a much wider range of budgets and ways to
allocate them. Router scores have also become more diverse, with softmax gating, sigmoid gating with
learned expert biases, and the square root of softplus \citep{deepseekv4}. These changes make fine-grained MoE a compelling setting for revisiting expert
redundancy: both the number of selections available to prune and the signals used to prune them have
evolved substantially.

\subsection{Dynamic expert pruning}
\label{sec:related:pruning}

Dynamic expert pruning retains the expert pool while reducing the subset executed at inference time. It
complements structural pruning, which permanently removes experts from the model altogether
\citep{muzio2024seer}.

Existing rules draw on different information to make this decision. \naee{} compares the second
expert's gate weight with the leading expert's and skips it when their ratio falls below a calibrated
threshold \citep{lu2024naee}. \diep{} incorporates similarity between expert outputs into this
comparison, making retention depend on both router preference and overlap between expert computations
\citep{diep2025}. Cumulative-probability routing, represented here by \dynr{}, moves from pairwise
comparisons to the selected distribution as a whole: it retains the shortest prefix whose cumulative
gate weight reaches a target \citep{huang2024dynamicmoe}. \ban{} adds a further allocation dimension by
combining an offline layer-sensitivity profile with an online token signal, distributing the expert
budget across both tokens and depth with fine-grained routers and long-form reasoning in view
\citep{ban}. EAC-MoE operates at a different granularity, restricting pruning to prefill and basing its
decisions on the traffic received by each expert \citep{eacmoe2025}.

These approaches provide increasingly rich information for allocating expert computation. Our study
examines how much that allocation contributes beyond the redundancy captured by simply lowering
$\ktop$. By comparing representative rules across fine-grained architectures, task families and matched
expert budgets, we establish when dynamic allocation improves on uniform truncation and where future
methods have the most room to advance.

\section{Experimental setup}
\label{sec:setup}

Our experiments use the routed expert budget as a common axis for redundancy, dynamic allocation
and pruning sensitivity. We combine broad coverage of fine-grained MoE architectures with
focused comparisons of model properties, and evaluate knowledge and reasoning across three task
families.

\subsection{Routing and expert budgets}
\label{sec:setup:routing}

An MoE layer contains $\nexp$ routed expert networks $f_1,\dots,f_{\nexp}$ and a router. Given a token
representation $\vx$, the router selects $\ktop$ experts with indices $e_1,\dots,e_{\ktop}$, ordered by
descending gate weights $w_1\ge\dots\ge w_{\ktop}$. The weights are normalised over the selected
experts, giving the layer output
\begin{equation}
\mathrm{MoE}(\vx) \;=\; \sum_{i=1}^{\ktop} w_i\, f_{e_i}(\vx)
\;+\; \sum_{j=1}^{n_s} f^{\mathrm{shared}}_{j}(\vx).
\label{eq:moe}
\end{equation}
Here, $n_s$ denotes the shared experts applied to every token. The native layer executes $\ktop+n_s$
expert networks per token: $\nexp$ buys capacity, $\ktop$ pays for it.

Pruning retains a subset of the selected experts. Let $\mathcal{S}(\vx,\ell)$ denote their ranks at
layer $\ell$, with $k(\vx,\ell)=|\mathcal{S}(\vx,\ell)|$. Renormalising the retained weights gives
\begin{equation}
\widetilde{\mathrm{MoE}}(\vx) \;=\; \sum_{i\in\mathcal{S}(\vx,\ell)} \tilde{w}_i\, f_{e_i}(\vx)
\;+\; \sum_{j=1}^{n_s} f^{\mathrm{shared}}_{j}(\vx),
\qquad
\tilde{w}_i \;=\; \frac{w_i}{\sum_{j\in\mathcal{S}(\vx,\ell)} w_j}.
\label{eq:pruned}
\end{equation}
Uniform truncation, denoted \fixedk{}, keeps the top $k$ experts for every token and layer. Dynamic
rules determine the retained experts from signals that vary across tokens and layers. Both reduce
routed expert execution and its associated data movement, while shared experts continue to process
every token. This common formulation lets us distinguish the effect of reducing the budget from the
benefit of allocating it dynamically.

\subsection{Models}
\label{sec:setup:models}

Our model selection emphasises architectural breadth rather than depth within any single family.
Table~\ref{tab:models} summarises the twelve checkpoints and their roles.

Nine checkpoints span nine architecture families \citep{linglite2507, gptoss,
deepseekv2, hy3, gemma4}, with routed expert pools of $32$--$512$ and native
selections of $4$--$10$ experts per token. They include models with and without shared experts, and
routers scored by softmax, sigmoid and the square root of softplus. This range lets us examine whether substantial routing redundancy
persists across different fine-grained designs. Four core models support both the uniform budget sweeps
(\S\ref{sec:redundancy}) and the comparison of dynamic rules (\S\ref{sec:dynamic}); MiniMax-M2.7,
DeepSeek-V2-Lite-Chat, Hy3, DeepSeek-V4-Flash and Gemma 4 26B-A4B extend the uniform sweeps to five
further architectures.

Three further Qwen3 checkpoints support the analysis of reasoning-mode post-training, parameter scale
and modality \citep{qwen3vl} (\S\ref{sec:axes}). All share a routed expert pool of $\nexp=128$ and a native selection of
$\ktop=8$, giving these comparisons a common budget axis. Together, the two groups connect routing
redundancy to the model properties that shape pruning sensitivity.

\begin{table}[t]
\centering
\small
\setlength{\tabcolsep}{2.15pt}
\renewcommand{\arraystretch}{1.12}
\caption{Models. Total and active are parameter counts; $\nexp$ is the routed experts per layer, $\ktop$
how many the router natively activates per token, and $n_s$ the shared experts every token passes
through.}
\label{tab:models}
\newcommand{\grouprow}[2]{%
  \rowcolor{groupfill}\multicolumn{7}{l}{%
    \rule{0pt}{11pt}\textcolor{groupink}{\textbf{#1}}%
    \hspace{0.9em}\textcolor{groupink}{\footnotesize #2}}\\[1pt]}
\begin{tabularx}{\textwidth}{>{\hsize=1.40\hsize}Lrrrrr>{\hsize=0.60\hsize}L}
\toprule
\textbf{Model} & \textbf{Total} & \textbf{Active} & $\bm{\nexp}$ & $\bm{\ktop}$ & $\bm{n_s}$
& \textbf{Role} \\
\midrule
\grouprow{Core models}{uniform sweeps (\S\ref{sec:redundancy}) and dynamic rules (\S\ref{sec:dynamic})}
Qwen3-30B-A3B-Instruct \citep{qwen3} & 30.5B & 3.3B  & 128 & 8  & 0 & reference model \\
Qwen3-Next-80B-A3B-Instruct \citep{qwen3next} & 80B   & 3B    & 512 & 10 & 1 & widest router \\
Ling-lite-1.5-2507 \citep{linglite2507} & 16.8B & 2.75B & 64  & 6  & 2 & shared experts \\
GPT-OSS-20B \citep{gptoss} & 21B   & 3.6B  & 32  & 4  & 0 & narrowest router \\
\addlinespace[3pt]
\grouprow{Additional architectures}{uniform sweeps only (\S\ref{sec:redundancy})}
MiniMax-M2.7 \citep{minimaxm2} & 229B  & 11B   & 256 & 8  & 0 & sigmoid router \\
DeepSeek-V2-Lite-Chat \citep{deepseekv2} & 15.7B & 2.4B  & 64  & 6  & 2 & earliest fine-grained \\
Hy3 \citep{hy3} & 295B  & 21B   & 192 & 8  & 1 & largest model \\
DeepSeek-V4-Flash-0731 \citep{deepseekv4} & 165B  & 11.5B & 256 & 6  & 1 & sqrt-softplus router \\
Gemma 4 26B-A4B \citep{gemma4} & 25.2B & 3.8B  & 128 & 8  & 1 & GELU experts \\
\addlinespace[3pt]
\grouprow{Single-factor checkpoints}{one property changed (\S\ref{sec:axes})}
Qwen3-30B-A3B-Thinking \citep{qwen3} & 30.5B & 3.3B  & 128 & 8  & 0 & reasoning mode \\
Qwen3-235B-A22B-Instruct \citep{qwen3} & 235B  & 22B   & 128 & 8  & 0 & parameter scale \\
Qwen3-VL-30B-A3B-Instruct \citep{qwen3vl} & 30B   & 3B    & 128 & 8  & 0 & modality \\
\bottomrule
\end{tabularx}
\vspace{-8pt}
\end{table}

\subsection{Tasks and evaluation}
\label{sec:setup:data}

The core evaluation contains eleven benchmarks organised into three task families, covering knowledge
mastery, multi-step reasoning and the application of knowledge across domains. Knowledge QA uses
likelihood scoring \citep{lmevalharness}; the two reasoning families are evaluated through generated
answers \citep{lighteval}.

\begin{flushitems}
\item \textbf{Knowledge QA.} ARC-Easy, ARC-Challenge \citep{arc}, WinoGrande \citep{winogrande} and
OpenBookQA \citep{openbookqa} assess the model's
command of scientific and commonsense knowledge. We score each candidate answer by its log-likelihood
under zero-shot prompting. This family measures how well knowledge survives a smaller budget.

\item \textbf{Mathematical and code reasoning.} AIME-24 \citep{aime24}, AIME-25 \citep{aime25},
MATH-500 \citep{hendrycksmath, math500}, GSM8K \citep{gsm8k} and LiveCodeBench-v6
\citep{livecodebench} test multi-step reasoning, from grade-school arithmetic to competition mathematics and
programming. Successful answers require carrying a derivation through to a solution or producing a
working program.

\item \textbf{Knowledge and general reasoning.} MMLU-Pro \citep{mmlupro} and GPQA-Diamond \citep{gpqa} assess broad
subject knowledge and its application to reasoning across domains. They extend the evaluation to
graduate-level science and other challenging problems, combining what a model knows with what it can
infer.
\end{flushitems}

The modality analysis in \S\ref{sec:axes} adds a nine-dataset multimodal suite and a text QA suite
expanded to nine datasets, both listed in Appendix~\ref{app:full:axes}. Answers range from short
factual responses to long derivations and complete programs. This breadth lets us examine whether pruning preserves both knowledge
and the ability to use it in a complete solution. We begin in \S\ref{sec:redundancy} by lowering the
expert count uniformly, establishing how much of the native selection these capabilities actually
require.

\section{How much routing does a fine-grained MoE actually need?}
\label{sec:redundancy}

How much of a fine-grained router's selection does inference actually need? We answer this by lowering
the expert count uniformly across tokens and layers, without changing the model parameters or the
router's ranking. This one-integer intervention isolates the redundancy already present in the model:
whatever quality it preserves at a reduced budget requires no dynamic allocation to recover. It also
establishes the baseline that a more elaborate pruning rule must improve upon.

\begin{figure}[t]
\centering
\includegraphics[width=\linewidth]{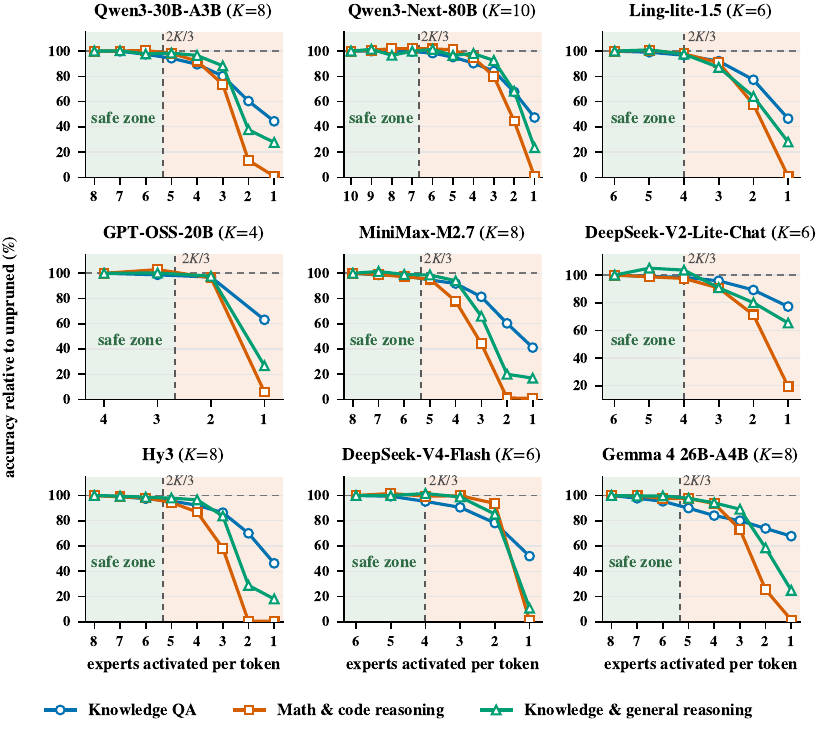}
\caption{Expert-selection redundancy under uniform truncation. Each panel is one model, with its own
native $\ktop$ on the horizontal axis and the budget tightening to the right; curves are the mean over
the datasets of a task family, normalised to that family's score on the unpruned model. The dashed
vertical line marks $2\ktop/3$.}
\label{fig:redundancy}
\end{figure}

\subsection{Two thirds of the selected experts are enough}
\label{sec:redundancy:curve}

Uniformly retaining $\lceil 2\ktop/3 \rceil$ experts preserves $98.8\%$ of unpruned performance,
averaged over nine models and three task families (Figure~\ref{fig:redundancy}). Six of the twenty-seven
model-family combinations match or exceed their unpruned scores, including mathematical and code
reasoning on both Qwen3 models and GPT-OSS-20B. Across architectures with native selections of four to
ten experts, substantial savings are therefore available through a one-integer change. Put plainly:
throw away roughly the lowest-weight third of the selected experts, and the model barely notices.

The gate-weighted mixture helps explain this tolerance (Equation~\ref{eq:pruned}). \fixedk{} removes
the router's lowest-weight choices and renormalises the remaining weights, concentrating the mixture on
its preferred experts. Removing a third of the selected experts therefore need not remove a third of
their contribution. The sweep reveals how little the full selection adds in this regime.

As the budget tightens, this spare capacity runs out. At three of eight experts, Qwen3-30B-A3B retains
$73.6\%$ of its mathematical and code reasoning score; at two of six, Ling-lite-1.5 retains $57.8\%$.
These intermediate budgets already incur substantial losses, even though the models retain much of
their ability to solve the tasks.

Further pruning brings a much sharper breakdown, and the same absolute expert count can mean radically
different things across models. Keeping two experts leaves Qwen3-30B-A3B at just $13.5\%$ on
mathematics and code, but GPT-OSS-20B at $96.3\%$: two experts are a quarter of the former's native
selection and half of the latter's. Reducing GPT-OSS-20B to one expert then sends its score to $6.0\%$.
The retained fraction of the native selection is thus essential to interpreting a pruning budget.

\subsection{Knowledge survives what reasoning does not}
\label{sec:redundancy:regimes}

Budget tightening also changes which capabilities are most resilient. Knowledge QA declines gradually
from the start, while the generative families initially hold steady and then fall sharply. On
Qwen3-30B-A3B, knowledge QA retains $97.5\%$ of its unpruned score at six experts, below both
generative families. At two experts, that ordering reverses: knowledge QA still retains $60.4\%$, while
mathematical and code reasoning falls to $13.5\%$. On eight of the nine models, mathematical and code
reasoning is the least robust family at the aggressive end.

This reversal separates retaining knowledge from sustaining a complete solution. Knowledge QA tests the
model's command of scientific and commonsense knowledge through short candidate answers. A derivation
or program must carry that knowledge through a sequence of dependent steps. Each generated step becomes
context for the next, allowing local disruptions to propagate through the rest of the answer. Long-form
generation therefore creates repeated opportunities for pruning errors to compound. Evaluating both
reveals a distinction that matters for deployment: a model can retain much of what it knows while
losing the ability to use it in an extended solution.

\subsection{From routing redundancy to inference speed}
\label{sec:speedup}

\begin{table}[t]
\centering
\small
\setlength{\tabcolsep}{5pt}
\caption{Measured speedup from uniform truncation at the tested reduced budgets. Throughput ratios are
relative to the same unpruned model on each backend, reported separately for prefill and decode. All runs
use batch size $4$, $1024$-token prompts and $256$ generated tokens.}
\label{tab:speed}
\begin{tabular}{l ccc @{\hspace{16pt}} cc @{\hspace{12pt}} cc}
\toprule
& \multicolumn{3}{c}{Operating point} & \multicolumn{2}{c}{vLLM}
& \multicolumn{2}{c}{HF Transformers} \\
\cmidrule(lr){2-4}\cmidrule(lr){5-6}\cmidrule(lr){7-8}
Model & $\ktop$ & $\kbar$ & $\rbudget$ & prefill & decode & prefill & decode \\
\midrule
Qwen3-30B-A3B  & 8  & 5 & 0.63 & 1.15 & 1.18 & 1.26 & 1.70 \\
Qwen3-Next-80B & 10 & 6 & 0.60 & 1.25 & 1.65 & 1.31 & 1.73 \\
Ling-lite-1.5  & 6  & 4 & 0.67 & 1.21 & 1.38 & 1.21 & 1.52 \\
GPT-OSS-20B    & 4  & 3 & 0.75 & 1.25 & 1.23 & 1.24 & 1.44 \\
\bottomrule
\end{tabular}
\end{table}

The redundancy translates into measurable inference gains. Across the four core models at the reduced
budgets listed in Table~\ref{tab:speed}, \fixedk{} improves decode throughput by $1.18$--$1.65\times$
on vLLM and $1.44$--$1.73\times$ on HF Transformers. Prefill also improves, by $1.15$--$1.25\times$ and
$1.21$--$1.31\times$, respectively. A simple reduction in the expert count therefore delivers speedups
in both a reference implementation and an optimised serving stack.

The backend difference reflects how much expert execution still costs. In our HF Transformers
implementation, experts run as separate unfused GEMMs, so truncation removes individual matrix
multiplications and their associated overhead. vLLM's fused MoE kernels already amortise part of that
cost, leaving a smaller share for pruning to recover. Even there, Qwen3-Next-80B gains $65\%$ in decode
throughput; it achieves the largest decode gain on both backends. The practical return comes from
reducing routed expert computation and its associated data movement, with its size determined by the
model and serving implementation. These figures belong to \fixedk{}, which keeps every token at the
same expert count and leaves the expert GEMMs uniformly shaped; a dynamic rule gives each token its own
$k$, which is harder to serve efficiently at a matched average budget.

\begin{takeaway}
\finding{Takeaway.}{Fine-grained expert selection is far more redundant than native budgets suggest: a
one-integer reduction to roughly two thirds of the selected experts preserves $98.8\%$ of performance
on average. Its limits emerge under aggressive pruning, where long-form reasoning deteriorates most
sharply. This is where dynamic allocation faces its decisive test: can redistributing the same expert
budget preserve the reasoning that uniform truncation loses (\S\ref{sec:dynamic})?}
\end{takeaway}

\section{When does dynamic allocation beat uniform pruning?}
\label{sec:dynamic}

Uniform pruning reveals how much routing is dispensable. Dynamic pruning asks when deciding where to
spend the remaining budget starts to matter. We compare four published rules with \fixedk{} on four
fine-grained MoEs at matched budget tiers. The result is a sharp change in value: adaptation adds
almost nothing at conservative budgets, but recovers up to $3.0$ points once pruning turns aggressive.

\subsection{Comparing allocation at matched budgets}
\label{sec:dynamic:rules}

We implement four allocation rules in a common router patch on vLLM. Their key difference is the
information used to decide which experts to retain. \naee{} thresholds an expert's gate weight relative
to the leading expert. \dynr{} retains the shortest ranked prefix reaching a target
cumulative probability mass, renormalised over the native selection. We
evaluate \diep{}'s similarity-adjusted skipping criterion, generalised to fine-grained
routing, together with a variant that damps the similarity term. \ban{} combines token-level routing
information with a calibrated layer-sensitivity profile. The rules and their calibration are detailed in Appendices~\ref{app:setup}
and~\ref{app:diep}.

To compare these different signals, we characterise each configuration by its measured average number
of active routed experts:
\begin{equation}
\kbar \;=\; \frac{1}{|\mathcal{T}|\,L}\sum_{\vx\in\mathcal{T}}\sum_{\ell=1}^{L} k(\vx,\ell),
\qquad \rbudget \;=\; \frac{\kbar}{\ktop} .
\label{eq:budget}
\end{equation}
Here, $\mathcal{T}$ contains the evaluated tokens and $L$ is the number of MoE layers. We measure
$\kbar$ inside the router during evaluation. Thresholds are calibrated separately for generation and
likelihood-scored QA, since the same setting can spend different budgets in the two regimes.
Table~\ref{tab:iso} reports the realised budgets and pairs \fixedk{} with the best eligible rule at each
tier, selected by its mean score over all eleven datasets. Configurations exceeding their target budget
by more than $10\%$ are excluded from this comparison.

\subsection{Dynamic allocation pays off under aggressive pruning}
\label{sec:dynamic:results}

\begin{table}[t]
\centering
\scriptsize
\setlength{\tabcolsep}{0.92pt}
\caption{Iso-budget comparison, per dataset. Each model is given two budgets, and under each the uniform
baseline is paired with the best of the four dynamic rules at that same budget, selected by the mean
over all eleven datasets (last column). The superscript is that rule's margin over the \fixedk{} row
directly above it. The budget beside a rule is its measured $\kbar$ for generative\,/\,QA, since the
knob is solved separately per regime (\S\ref{sec:dynamic:rules}); rules spending more than $10\%$ over
the tier are excluded.}
\label{tab:iso}
\begin{tabularx}{\linewidth}{L cccc ccccc cc c}
\toprule
& \multicolumn{4}{c}{Knowledge QA}
& \multicolumn{5}{c}{Math \& code reasoning}
& \multicolumn{2}{c}{Knowledge \& general} & \\
\cmidrule(lr){2-5}\cmidrule(lr){6-10}\cmidrule(lr){11-12}
& ARC-c & ARC-e & WG & OBQA
& MATH & AIME24 & AIME25 & GSM8K & LCB
& MMLU-P & GPQA-D & Avg \\
\midrule
\multicolumn{13}{l}{\textbf{Qwen3-30B-A3B}, natively $\ktop{=}8$}\\
\addlinespace[3pt]
\rowcolor{unprunedfill}\quad unpruned, $k{=}8$ & 60.8 & 85.0 & 73.2 & 32.0 & 89.6 & 74.1 & 61.4 & 94.4 & 41.1 & 74.2 & 56.6 & 67.49 \\
\rowcolor{consfill}\multicolumn{13}{l}{\quad\textit{Conservative budget}, $\kbar\!\approx\!5$ ($\rbudget{=}0.62$)}\\
\rowcolor{consfill}\qquad \fixedk{}, $k{=}5$ & 54.9 & 81.4 & 69.5 & 31.0 & 91.0 & 72.1 & 55.4 & 93.9 & 41.7 & 73.8 & 55.0 & 65.43 \\
\rowcolor{consfill}\qquad \textbf{\dynr{}}, $4.91/5.09$ & 56.5 & 81.4 & 71.6 & 31.0 & 89.6 & 73.2 & 55.2 & 94.2 & 41.7 & 73.2 & 55.1 & \textbf{65.70}$^{+0.27}$ \\
\rowcolor{aggrfill}\multicolumn{13}{l}{\quad\textit{Aggressive budget}, $\kbar\!\approx\!3$ ($\rbudget{=}0.38$)}\\
\rowcolor{aggrfill}\qquad \fixedk{}, $k{=}3$ & 44.6 & 71.8 & 58.0 & 27.0 & 80.2 & 40.0 & 31.1 & 88.5 & 25.7 & 67.0 & 48.5 & 52.95 \\
\rowcolor{aggrfill}\qquad \textbf{\ban{}}, $3.14/3.08$ & 46.1 & 72.8 & 61.5 & 25.8 & 88.0 & 42.9 & 34.0 & 90.4 & 28.6 & 68.8 & 50.5 & \textbf{55.40}$^{+2.45}$ \\
\midrule
\multicolumn{13}{l}{\textbf{Qwen3-Next-80B}, natively $\ktop{=}10$}\\
\addlinespace[3pt]
\rowcolor{unprunedfill}\quad unpruned, $k{=}10$ & 63.1 & 86.9 & 76.2 & 34.2 & 88.4 & 79.7 & 67.1 & 95.2 & 53.1 & 82.1 & 74.2 & 72.75 \\
\rowcolor{consfill}\multicolumn{13}{l}{\quad\textit{Conservative budget}, $\kbar\!\approx\!6$ ($\rbudget{=}0.60$)}\\
\rowcolor{consfill}\qquad \fixedk{}, $k{=}6$ & 63.1 & 85.6 & 73.9 & 33.6 & 90.8 & 82.2 & 66.6 & 95.7 & 54.9 & 82.1 & 76.3 & 73.16 \\
\rowcolor{consfill}\qquad \textbf{\ban{}}, $6.18/6.08$ & 62.4 & 85.7 & 74.1 & 32.4 & 89.2 & 82.7 & 69.7 & 95.3 & 54.9 & 82.2 & 74.2 & 72.98$^{-0.18}$ \\
\rowcolor{aggrfill}\multicolumn{13}{l}{\quad\textit{Aggressive budget}, $\kbar\!\approx\!4$ ($\rbudget{=}0.40$)}\\
\rowcolor{aggrfill}\qquad \fixedk{}, $k{=}4$ & 55.9 & 82.2 & 67.2 & 30.2 & 88.0 & 74.8 & 58.8 & 93.1 & 49.1 & 80.7 & 72.7 & 68.43 \\
\rowcolor{aggrfill}\qquad \textbf{\ban{}}, $4.27/4.18$ & 59.8 & 83.5 & 67.8 & 32.8 & 88.6 & 79.4 & 64.6 & 94.4 & 57.1 & 81.3 & 73.2 & \textbf{71.14}$^{+2.71}$ \\
\midrule
\multicolumn{13}{l}{\textbf{Ling-lite-1.5}, natively $\ktop{=}6$}\\
\addlinespace[3pt]
\rowcolor{unprunedfill}\quad unpruned, $k{=}6$ & 60.2 & 82.3 & 70.3 & 31.6 & 89.2 & 40.9 & 28.1 & 91.4 & 34.3 & 69.4 & 56.6 & 59.48 \\
\rowcolor{consfill}\multicolumn{13}{l}{\quad\textit{Conservative budget}, $\kbar\!\approx\!4$ ($\rbudget{=}0.67$)}\\
\rowcolor{consfill}\qquad \fixedk{}, $k{=}4$ & 58.3 & 81.2 & 66.1 & 30.8 & 89.0 & 38.7 & 26.0 & 90.5 & 33.7 & 67.4 & 55.6 & 57.94 \\
\rowcolor{consfill}\qquad \textbf{\dynr{}}, $4.17/4.22$ & 58.4 & 82.0 & 68.4 & 30.4 & 90.2 & 40.1 & 26.8 & 90.5 & 32.0 & 67.3 & 58.6 & \textbf{58.61}$^{+0.67}$ \\
\rowcolor{aggrfill}\multicolumn{13}{l}{\quad\textit{Aggressive budget}, $\kbar\!\approx\!2$ ($\rbudget{=}0.33$)}\\
\rowcolor{aggrfill}\qquad \fixedk{}, $k{=}2$ & 38.9 & 69.6 & 57.4 & 23.0 & 70.4 & 5.0 & 8.4 & 72.4 & 8.0 & 42.3 & 38.4 & 39.44 \\
\rowcolor{aggrfill}\qquad \textbf{\ban{}}, $2.05$ & 42.2 & 71.2 & 57.1 & 24.6 & 70.4 & 10.2 & 11.5 & 75.1 & 14.9 & 47.5 & 41.9 & \textbf{42.42}$^{+2.98}$ \\
\midrule
\multicolumn{13}{l}{\textbf{GPT-OSS-20B}, natively $\ktop{=}4$}\\
\addlinespace[3pt]
\rowcolor{unprunedfill}\quad unpruned, $k{=}4$ & 45.3 & 77.5 & 66.9 & 27.2 & 90.6 & 73.7 & 71.7 & 86.2 & 58.3 & 74.3 & 64.6 & 66.94 \\
\rowcolor{consfill}\multicolumn{13}{l}{\quad\textit{Conservative budget}, $\kbar\!\approx\!3$ ($\rbudget{=}0.75$)}\\
\rowcolor{consfill}\qquad \fixedk{}, $k{=}3$ & 44.8 & 76.7 & 65.4 & 27.4 & 91.0 & 77.4 & 74.5 & 87.0 & 61.7 & 73.6 & 65.7 & 67.75 \\
\rowcolor{consfill}\qquad \textbf{\dynr{}}, $2.97$ & 43.4 & 77.8 & 65.1 & 28.2 & 90.0 & 77.0 & 72.4 & 86.6 & 62.3 & 73.5 & 63.1 & 67.21$^{-0.53}$ \\
\rowcolor{aggrfill}\multicolumn{13}{l}{\quad\textit{Aggressive budget}, $\kbar\!\approx\!2$ ($\rbudget{=}0.50$)}\\
\rowcolor{aggrfill}\qquad \fixedk{}, $k{=}2$ & 44.3 & 75.9 & 64.1 & 26.0 & 86.6 & 73.9 & 69.4 & 88.7 & 48.0 & 70.5 & 65.2 & 64.78 \\
\rowcolor{aggrfill}\qquad \textbf{\dynr{}}, $2.07$ & 42.3 & 73.6 & 63.1 & 26.1 & 89.9 & 73.7 & 71.2 & 85.6 & 62.5 & 70.8 & 63.6 & \textbf{65.69}$^{+0.90}$ \\
\bottomrule
\end{tabularx}
\end{table}

At conservative budgets, even the best dynamic rule offers little advantage. Across the four models, its
margin over \fixedk{} ranges from $-0.53$ to $+0.67$ points, averaging just $+0.06$. The redundancy
found in \S\ref{sec:redundancy} therefore leaves little room for a more elaborate allocation: giving
every token the same reduced expert count already matches the best tested rules.

Tightening the budget changes this result on every model. At target budgets retaining between one third
and one half of the native selection, the best rule gains $2.45$ points on Qwen3-30B-A3B, $2.71$ on
Qwen3-Next-80B, $2.98$ on Ling-lite-1.5 and $0.90$ on GPT-OSS-20B. The average gain rises to $2.26$
points. Allocation now recovers accuracy that uniform truncation loses at the corresponding budget.
This is where dynamic pruning earns its place: once the expert budget is tight enough to damage
performance, deciding where to spend it becomes consequential.

The budget also changes which allocation rule performs best. On Qwen3-30B-A3B, \dynr{} has the highest
conservative-budget mean, whereas \ban{} wins under aggressive pruning. \ban{} achieves the best
aggressive-budget result on three of the four models, pointing to calibrated layer sensitivity as a
promising signal for allocating scarce expert computation. Together, these results make the budget part
of the method comparison: an evaluation at one operating point can miss both the gains from adaptation
and the rule that delivers them.

\subsection{Generative tasks reveal the value of reallocation}
\label{sec:dynamic:tasks}

The recovery is concentrated on generative tasks. At Qwen3-30B-A3B's aggressive budget,
\ban{} improves the mean over seven generative datasets by $3.17$ points, compared with $1.20$ on the
four Knowledge QA datasets. Within that mean, MATH-500 rises from $80.2$ to $88.0$ and
LiveCodeBench from $25.7$ to $28.6$. The same pattern is striking on GPT-OSS-20B: \dynr{} raises
LiveCodeBench from $48.0$ to $62.5$, a $14.5$-point gain within a suite-wide improvement of $0.90$
points, since its other datasets already sit close to their unpruned scores at this budget.

These results connect the value of allocation directly to the task split in \S\ref{sec:redundancy}.
Generation exposes a sharper loss under uniform pruning and a larger recovery when the remaining budget
is distributed dynamically. Knowledge QA alone captures only part of that change. A pruning evaluation
dominated by likelihood-scored answers can therefore understate both the cost of reducing the expert
count and the benefit of adapting it. Requiring the model to produce a complete derivation, program or
answer reveals where the allocation decision has the greatest effect.

\begin{takeaway}
\finding{Takeaway.}{Dynamic allocation earns its place under aggressive pruning. At conservative
budgets, even the best tested rule remains within a fraction of a point of uniform truncation; at
aggressive budgets, the gains reach $3.0$ points and concentrate on generative tasks. Budget
determines when adaptation pays off, while the evaluation tasks determine how much of that value is
visible.}
\end{takeaway}

\section{What governs a fine-grained MoE's sensitivity to expert pruning?}
\label{sec:axes}

The preceding sections show that both the cost of pruning and the value of dynamic allocation depend on
the budget. We now ask which model properties govern the ability to tolerate a tighter budget. We
compare reasoning-mode post-training, parameter scale and modality through three pairs drawn from four
Qwen3 checkpoints. All have $128$ experts and natively select eight; each curve is normalised to its own
unpruned performance. Across these comparisons, the main differences emerge under aggressive pruning.

\begin{figure}[t]
\centering
\includegraphics[width=\linewidth]{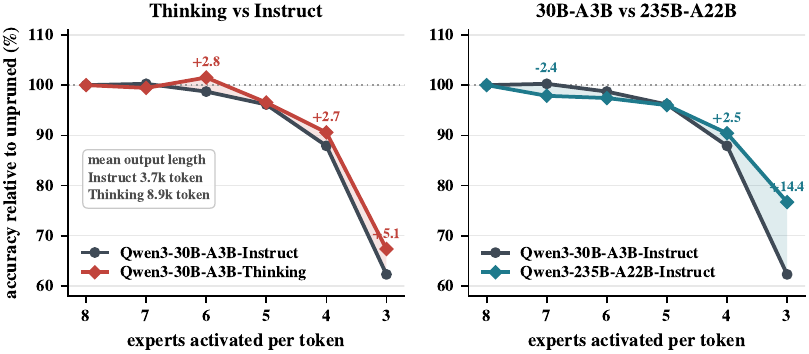}
\caption{Two matched pairs under uniform truncation, each curve normalised to that model's own unpruned
score. Curves are the mean over the four hardest generative datasets, AIME-24, AIME-25, LiveCodeBench
and GPQA-Diamond; annotations give the gap in points where it exceeds two. The axis stops at
three experts because below it both members of each pair have collapsed on these tasks, so the remaining
differences are not interpretable.}
\label{fig:axes}
\end{figure}

\subsection{Thinking preserves more generative capability}
\label{sec:axes:reasoning}

Reasoning-mode post-training preserves more generative capability as the expert budget tightens. We
compare Qwen3-30B-A3B-Thinking with its Instruct counterpart on the four challenging generative datasets
in Figure~\ref{fig:axes}: AIME-24, AIME-25, LiveCodeBench and GPQA-Diamond. Their performance
retention remains close from eight to five experts. At four experts, the thinking model leads by $2.7$
percentage points; at three, its advantage grows to $5.1$ points.

The task breakdown in Appendix~\ref{app:full:axes} locates this advantage in generation, with little
difference on Knowledge QA. The thinking model also produces longer outputs, averaging $8.9$k tokens per
task against Instruct's $3.7$k. It thus combines longer generation with better performance retention
under aggressive pruning. The benefit of reasoning-mode post-training becomes visible precisely on the
capabilities that \S\ref{sec:redundancy} found most vulnerable to reducing the expert count.

\subsection{Scale matters most under aggressive pruning}
\label{sec:axes:scale}

Parameter scale produces an even larger separation as pruning deepens. On the same four generative
datasets, Qwen3-235B-A22B-Instruct holds no advantage over Qwen3-30B-A3B-Instruct
from seven to five experts. At four experts, the larger model moves $2.5$ points ahead. At
three, it retains $76.7\%$ of its unpruned performance while the smaller model retains $62.3\%$, opening
a $14.4$-point gap.

As with reasoning mode, the broader task breakdown places the advantage in mathematical, code and
general reasoning, with no corresponding gain on Knowledge QA. The extra scale preserves substantially
more generative performance once the budget becomes restrictive. A comparison confined to conservative
pruning would reveal little of this difference: the benefit emerges when retaining the same number of
experts leaves the smaller model substantially more impaired.

\begin{figure}[t]
\centering
\includegraphics[width=\linewidth]{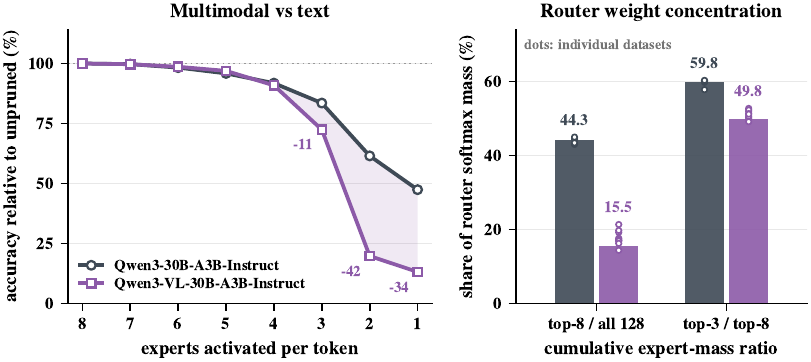}
\caption{Modality, measured on the scores and on the router. Left: uniform truncation on nine datasets a
side, the multimodal suite against nine likelihood-scored QA datasets on the text model, each curve
normalised to its own unpruned score; annotations give the gap in points. Right: the router's full
$128$-way softmax, recorded before top-$k$ selection over all $48$ MoE layers and weighted by token
count. Dots are the individual datasets.}
\label{fig:modality}
\end{figure}

\subsection{Multimodal models are more sensitive to aggressive pruning}
\label{sec:axes:modality}

The multimodal comparison reveals a sharper decline at tight budgets. Figure~\ref{fig:modality} compares
Qwen3-VL-30B-A3B-Instruct with Qwen3-30B-A3B-Instruct on nine datasets each: multimodal tasks for the
former and likelihood-scored QA for the latter. Their normalised curves remain within one percentage
point from eight to four experts. At three experts, the multimodal model falls $11$ points behind; at
two, the gap widens to $42$ points. The similar response to conservative pruning gives way to sharply
different losses as the budget tightens.

Routing statistics reveal a corresponding difference in weight concentration. We record each router's
full $128$-way softmax at every one of the $48$ MoE layers, before top-$\ktop$ selection, while the two
models run the same suites as above, and aggregate the distributions over evaluated tokens. Both models
are measured on prompt tokens, the comparable stage across the pair. Within the selected eight
experts, the leading three hold $49.8\%$ of the multimodal model's routed weight, compared with $59.8\%$
for the text model. Truncating to three therefore discards $50.2\%$ and $40.2\%$ of the selected weight,
respectively, before renormalisation. The same retained expert count removes a larger share of routed
weight in the multimodal model.

This difference extends to the full routing distribution: the selected eight experts hold $15.5\%$ of
the $128$-way softmax mass in the multimodal model and $44.3\%$ in the text model. Per-dataset values do
not overlap on either concentration measure, and the direction is consistent across all $48$ MoE layers.
The flatter routing distribution helps explain why aggressive truncation is more disruptive in the
multimodal evaluation.

\begin{takeaway}
\finding{Takeaway.}{Model properties matter most once pruning becomes aggressive. Reasoning-mode
post-training and parameter scale preserve more generative capability, while the multimodal model loses
performance more sharply and distributes routing weight more broadly. The common tolerance at
conservative budgets gives way to distinct performance trajectories, making model properties central to
deciding how far pruning can go.}
\end{takeaway}

\section{Implications for dynamic expert pruning}
\label{sec:guidelines}

Our results distinguish two sources of gains from expert pruning: redundancy that uniform truncation can
remove, and additional capability that dynamic allocation preserves as budgets tighten. This distinction
connects the practical choice of a pruning budget to the evaluation and design of allocation rules.
Uniform pruning establishes how far expert computation can be reduced; its remaining performance losses
reveal where better allocation has something to recover. Building on these findings, we organise our
recommendations into three steps: establishing an operating budget with uniform truncation, measuring
what a dynamic rule adds on top of it, and designing allocation for the tight budgets where that
addition matters.

\subsection{Establishing the budget with uniform pruning}
\label{sec:guidelines:budget}

The broad tolerance observed in \S\ref{sec:redundancy} makes \textbf{uniform truncation the natural
starting point for budget selection}. Retaining $\lceil 2\ktop/3 \rceil$ experts provides a useful
initial operating point across the architectures studied, with the reduction controlled by a single
routing parameter. Progressively lowering that count then reveals how much further computation can be
reduced and where the model begins to lose capability. The resulting curve establishes what is already
achievable before introducing an adaptive allocation rule.

Where to stop on this curve depends on the capabilities the application needs to retain. For workloads
involving derivations, programs or open-ended answers, the sweep should track complete generations,
which expose the sharper pruning losses seen in \S\ref{sec:redundancy} and \S\ref{sec:dynamic}. The
model comparisons in \S\ref{sec:axes} help interpret how far the curve can extend: larger and thinking
checkpoints retain more generative capability under aggressive pruning, while the multimodal checkpoint
loses performance more sharply. A budget that preserves one checkpoint's performance can therefore sit
well beyond another's useful operating range.

\subsection{Evaluating the added value of dynamic allocation}
\label{sec:guidelines:evaluate}

Once a target budget pushes uniform truncation into noticeable degradation, \textbf{the relevant gain is
the capability recovered by reallocating that budget}. \S\ref{sec:dynamic} places the clearest gains
under aggressive pruning and on generative tasks. Measuring the margin over uniform truncation across
budget levels reveals where adaptation becomes useful and how the choice of rule changes. \ban{} and
\dynr{} provide useful starting candidates, supplying most of the winning dynamic configurations in
Table~\ref{tab:iso}.

The common basis for this comparison is the realised average number of activated experts, measured over
evaluation tokens and MoE layers. A threshold acquires its budget meaning through the routing
distributions it acts on, so that meaning can change with the model and workload. Remeasuring expert use
when transferring a setting, and recalibrating when it drifts from the target, keeps the comparison
focused on how effectively each rule allocates the available computation.

\subsection{Designing allocation for tight budgets}
\label{sec:guidelines:design}

The aggressive-budget results also point to \textbf{layer sensitivity as a promising complement to
token-level routing signals}: the strongest results under aggressive pruning come from distributing the
budget across depth as well as across tokens. The modality comparison in \S\ref{sec:axes} further shows why routing signals need to be
interpreted in the context of the target model. Retaining the same number of experts preserves different
shares of routing weight in the text and multimodal checkpoints. Together, these observations motivate
allocation rules that combine a token's routing distribution with the layer's measured response to
expert removal.

How that signal is parameterised then determines whether a calibrated setting survives a change of
architecture. Expressed as a ratio to the top-1 gate weight, a threshold inherits a bar that moves with
the width of the router: the leading expert holds most of the distribution on a top-2 router and a small
share of it on a top-10 router, so the same coefficient defines a different rule on each
(Appendix~\ref{app:diep}). Cumulative retained mass and per-rank statistics keep their meaning across
router widths, allowing a setting calibrated on one fine-grained model to carry to another without being
re-derived.

As for the inference gains actually realised, the practical value of these quality improvements depends
on the cost of executing the allocation that produces them. Efficient kernel support is needed to execute the selected token--expert assignments
while keeping allocation and execution overhead low. This allows the reduction in expert computation
from native routing to translate into practical inference savings. Evaluating latency or throughput
alongside retained quality then completes the comparison with uniform truncation, showing what the more
effective allocation delivers at its actual serving cost.

\section{Conclusion}
\label{sec:conclusion}

Our study of dynamic expert pruning in fine-grained MoEs, spanning twelve checkpoints across nine
architecture families, shows that routing is far more redundant than the field's operating points
assume. In the nine-model uniform sweep, retaining $\lceil 2\ktop/3 \rceil$ of the natively selected
experts preserves $98.8\%$ of baseline performance on average. Uniform pruning also achieves
$1.2$--$1.7\times$ decoding throughput across the tested serving configurations. A substantial reduction
in expert computation is therefore available by changing a single routing parameter.

The value of dynamic allocation emerges when pruning becomes aggressive. At conservative budgets, the
best tested rules offer only marginal gains over uniform truncation; at aggressive budgets, they recover
up to $3.0$ points, with the winning rule changing across models and budgets. Generative tasks reveal
this transition most clearly: they suffer sharper losses from expert removal and benefit more from
reallocating the remaining computation.

The same shift towards aggressive pruning also makes model properties consequential: larger and thinking
models retain more capability, whereas the multimodal checkpoint is more sensitive. These findings
establish uniform truncation as the essential reference for dynamic expert pruning. The clearest
opportunity for new allocation methods lies in preserving the generative capabilities that uniform
truncation begins to lose.

\bibliography{refs}

@inproceedings{shazeer2017outrageously,
  title     = {Outrageously Large Neural Networks: The Sparsely-Gated Mixture-of-Experts Layer},
  author    = {Shazeer, Noam and Mirhoseini, Azalia and Maziarz, Krzysztof and Davis, Andy and
               Le, Quoc and Hinton, Geoffrey and Dean, Jeff},
  booktitle = {International Conference on Learning Representations},
  year      = {2017}
}

@inproceedings{lepikhin2021gshard,
  title     = {{GShard}: Scaling Giant Models with Conditional Computation and Automatic Sharding},
  author    = {Lepikhin, Dmitry and Lee, HyoukJoong and Xu, Yuanzhong and Chen, Dehao and
               Firat, Orhan and Huang, Yanping and Krikun, Maxim and Shazeer, Noam and
               Chen, Zhifeng},
  booktitle = {International Conference on Learning Representations},
  year      = {2021}
}

@article{fedus2022switch,
  title   = {Switch Transformers: Scaling to Trillion Parameter Models with Simple and Efficient Sparsity},
  author  = {Fedus, William and Zoph, Barret and Shazeer, Noam},
  journal = {Journal of Machine Learning Research},
  volume  = {23},
  number  = {120},
  pages   = {1--39},
  year    = {2022},
  url     = {https://www.jmlr.org/papers/v23/21-0998.html}
}

@article{jiang2024mixtral,
  title   = {Mixtral of Experts},
  author  = {Jiang, Albert Q. and Sablayrolles, Alexandre and Roux, Antoine and Mensch, Arthur and
             Savary, Blanche and Bamford, Chris and Chaplot, Devendra Singh and
             de las Casas, Diego and Bou Hanna, Emma and Bressand, Florian and Lengyel, Gianna and
             Bour, Guillaume and Lample, Guillaume and Lavaud, L{\'e}lio Renard and
             Saulnier, Lucile and Lachaux, Marie-Anne and Stock, Pierre and
             Subramanian, Sandeep and Yang, Sophia and Antoniak, Szymon and Le Scao, Teven and
             Gervet, Th{\'e}ophile and Lavril, Thibaut and Wang, Thomas and Lacroix, Timoth{\'e}e and
             El Sayed, William},
  journal = {arXiv preprint arXiv:2401.04088},
  year    = {2024}
}

@inproceedings{dai2024deepseekmoe,
  title     = {{DeepSeekMoE}: Towards Ultimate Expert Specialization in Mixture-of-Experts Language Models},
  author    = {Dai, Damai and Deng, Chengqi and Zhao, Chenggang and Xu, R.X. and Gao, Huazuo and
               Chen, Deli and Li, Jiashi and Zeng, Wangding and Yu, Xingkai and Wu, Y. and
               Xie, Zhenda and Li, Y.K. and Huang, Panpan and Luo, Fuli and Ruan, Chong and
               Sui, Zhifang and Liang, Wenfeng},
  booktitle = {Proceedings of the 62nd Annual Meeting of the Association for Computational
               Linguistics (Volume 1: Long Papers)},
  pages     = {1280--1297},
  address   = {Bangkok, Thailand},
  publisher = {Association for Computational Linguistics},
  month     = aug,
  year      = {2024},
  doi       = {10.18653/v1/2024.acl-long.70}
}

@article{qwen3,
  title   = {{Qwen3} Technical Report},
  author  = {{Qwen Team}},
  journal = {arXiv preprint arXiv:2505.09388},
  year    = {2025}
}

@misc{qwen3next,
  title        = {{Qwen3-Next-80B-A3B-Instruct}},
  author       = {{Qwen Team}},
  year         = {2025},
  howpublished = {\url{https://huggingface.co/Qwen/Qwen3-Next-80B-A3B-Instruct}}
}

@article{glm5,
  title   = {{GLM-5}: from Vibe Coding to Agentic Engineering},
  author  = {{GLM-5 Team}},
  journal = {arXiv preprint arXiv:2602.15763},
  year    = {2026}
}

@article{minimaxm2,
  title   = {The {MiniMax-M2} Series: Mini Activations Unleashing Max Real-World Intelligence},
  author  = {{MiniMax}},
  journal = {arXiv preprint arXiv:2605.26494},
  year    = {2026}
}

@article{gemma4,
  title   = {{Gemma 4} Technical Report},
  author  = {{Gemma Team}},
  journal = {arXiv preprint arXiv:2607.02770},
  year    = {2026}
}

@misc{hy3,
  title        = {{Hy3}},
  author       = {{Tencent Hy Team}},
  year         = {2026},
  howpublished = {\url{https://huggingface.co/tencent/Hy3}}
}

@article{deepseekv4,
  title   = {{DeepSeek-V4}: Towards Highly Efficient Million-Token Context Intelligence},
  author  = {{DeepSeek-AI}},
  journal = {arXiv preprint arXiv:2606.19348},
  year    = {2026}
}

@article{kimik3,
  title   = {{Kimi K3}: Open Frontier Intelligence},
  author  = {{Kimi Team}},
  journal = {arXiv preprint arXiv:2607.24653},
  year    = {2026}
}

@inproceedings{lu2024naee,
  title     = {Not All Experts are Equal: Efficient Expert Pruning and Skipping for
               Mixture-of-Experts Large Language Models},
  author    = {Lu, Xudong and Liu, Qi and Xu, Yuhui and Zhou, Aojun and Huang, Siyuan and
               Zhang, Bo and Yan, Junchi and Li, Hongsheng},
  booktitle = {Proceedings of the 62nd Annual Meeting of the Association for Computational
               Linguistics (Volume 1: Long Papers)},
  pages     = {6159--6172},
  address   = {Bangkok, Thailand},
  publisher = {Association for Computational Linguistics},
  month     = aug,
  year      = {2024},
  doi       = {10.18653/v1/2024.acl-long.334}
}

@article{muzio2024seer,
  title   = {{SEER-MoE}: Sparse Expert Efficiency through Regularization for Mixture-of-Experts},
  author  = {Muzio, Alexandre and Sun, Alex and He, Churan},
  journal = {arXiv preprint arXiv:2404.05089},
  year    = {2024}
}

@inproceedings{diep2025,
  title     = {{DiEP}: Adaptive Mixture-of-Experts Compression through Differentiable Expert Pruning},
  author    = {Bai, Sikai and Li, Haoxi and Zhang, Jie and Hong, Zicong and Guo, Song},
  booktitle = {Advances in Neural Information Processing Systems},
  volume    = {38},
  pages     = {56090--56115},
  publisher = {Curran Associates, Inc.},
  year      = {2025},
  doi       = {10.52202/085713-1878}
}

@inproceedings{huang2024dynamicmoe,
  title     = {Harder Task Needs More Experts: Dynamic Routing in {MoE} Models},
  author    = {Huang, Quzhe and An, Zhenwei and Zhuang, Nan and Tao, Mingxu and Zhang, Chen and
               Jin, Yang and Xu, Kun and Xu, Kun and Chen, Liwei and Huang, Songfang and
               Feng, Yansong},
  booktitle = {Proceedings of the 62nd Annual Meeting of the Association for Computational
               Linguistics (Volume 1: Long Papers)},
  pages     = {12883--12895},
  address   = {Bangkok, Thailand},
  publisher = {Association for Computational Linguistics},
  month     = aug,
  year      = {2024},
  doi       = {10.18653/v1/2024.acl-long.696}
}

@inproceedings{ban,
  title     = {{Ban\&Pick}: Enhancing Performance and Efficiency of {MoE-LLMs} via Smarter Routing},
  author    = {Chen, Yuanteng and Wang, Peisong and Shao, Yuantian and Zeng, Nanxin and
               Xu, Chang and Cheng, Jian},
  booktitle = {Proceedings of the 2026 Conference on Empirical Methods in Natural Language
               Processing},
  year      = {2026},
  note      = {To appear. arXiv:2509.06346},
  url       = {https://arxiv.org/abs/2509.06346}
}

@inproceedings{eacmoe2025,
  title     = {{EAC-MoE}: Expert-Selection Aware Compressor for Mixture-of-Experts Large
               Language Models},
  author    = {Chen, Yuanteng and Shao, Yuantian and Wang, Peisong and Cheng, Jian},
  booktitle = {Proceedings of the 63rd Annual Meeting of the Association for Computational
               Linguistics (Volume 1: Long Papers)},
  pages     = {12942--12963},
  address   = {Vienna, Austria},
  publisher = {Association for Computational Linguistics},
  month     = jul,
  year      = {2025},
  doi       = {10.18653/v1/2025.acl-long.633}
}

@article{gptoss,
  title   = {{gpt-oss-120b} \& {gpt-oss-20b} Model Card},
  author  = {{OpenAI}},
  journal = {arXiv preprint arXiv:2508.10925},
  year    = {2025}
}

@article{deepseekv2,
  title   = {{DeepSeek-V2}: A Strong, Economical, and Efficient Mixture-of-Experts Language Model},
  author  = {{DeepSeek-AI}},
  journal = {arXiv preprint arXiv:2405.04434},
  year    = {2024}
}

@article{qwen3vl,
  title   = {{Qwen3-VL} Technical Report},
  author  = {Bai, Shuai and Cai, Yuxuan and Chen, Ruizhe and Chen, Keqin and Chen, Xionghui and Cheng, Zesen and Deng, Lianghao and Ding, Wei and Gao, Chang and Ge, Chunjiang and Ge, Wenbin and Guo, Zhifang and Huang, Qidong and Huang, Jie and Huang, Fei and Hui, Binyuan and Jiang, Shutong and Li, Zhaohai and Li, Mingsheng and Li, Mei and Li, Kaixin and Lin, Zicheng and Lin, Junyang and Liu, Xuejing and Liu, Jiawei and Liu, Chenglong and Liu, Yang and Liu, Dayiheng and Liu, Shixuan and Lu, Dunjie and Luo, Ruilin and Lv, Chenxu and Men, Rui and Meng, Lingchen and Ren, Xuancheng and Ren, Xingzhang and Song, Sibo and Sun, Yuchong and Tang, Jun and Tu, Jianhong and Wan, Jianqiang and Wang, Peng and Wang, Pengfei and Wang, Qiuyue and Wang, Yuxuan and Xie, Tianbao and Xu, Yiheng and Xu, Haiyang and Xu, Jin and Yang, Zhibo and Yang, Mingkun and Yang, Jianxin and Yang, An and Yu, Bowen and Zhang, Fei and Zhang, Hang and Zhang, Xi and Zheng, Bo and Zhong, Humen and Zhou, Jingren and Zhou, Fan and Zhou, Jing and Zhu, Yuanzhi and Zhu, Ke},
  journal = {arXiv preprint arXiv:2511.21631},
  year    = {2025}
}

@article{arc,
  title   = {Think you have Solved Question Answering? Try {ARC}, the {AI2} Reasoning Challenge},
  author  = {Clark, Peter and Cowhey, Isaac and Etzioni, Oren and Khot, Tushar and Sabharwal, Ashish and Schoenick, Carissa and Tafjord, Oyvind},
  journal = {arXiv preprint arXiv:1803.05457},
  year    = {2018}
}

@article{winogrande,
  title   = {{WinoGrande}: An Adversarial Winograd Schema Challenge at Scale},
  author  = {Sakaguchi, Keisuke and Le Bras, Ronan and Bhagavatula, Chandra and Choi, Yejin},
  journal = {arXiv preprint arXiv:1907.10641},
  year    = {2019}
}

@article{openbookqa,
  title   = {Can a Suit of Armor Conduct Electricity? A New Dataset for Open Book Question Answering},
  author  = {Mihaylov, Todor and Clark, Peter and Khot, Tushar and Sabharwal, Ashish},
  journal = {arXiv preprint arXiv:1809.02789},
  year    = {2018}
}

@article{hendrycksmath,
  title   = {Measuring Mathematical Problem Solving With the {MATH} Dataset},
  author  = {Hendrycks, Dan and Burns, Collin and Kadavath, Saurav and Arora, Akul and Basart, Steven and Tang, Eric and Song, Dawn and Steinhardt, Jacob},
  journal = {arXiv preprint arXiv:2103.03874},
  year    = {2021}
}

@article{math500,
  title   = {Let's Verify Step by Step},
  author  = {Lightman, Hunter and Kosaraju, Vineet and Burda, Yura and Edwards, Harri and Baker, Bowen and Lee, Teddy and Leike, Jan and Schulman, John and Sutskever, Ilya and Cobbe, Karl},
  journal = {arXiv preprint arXiv:2305.20050},
  year    = {2023}
}

@article{gsm8k,
  title   = {Training Verifiers to Solve Math Word Problems},
  author  = {Cobbe, Karl and Kosaraju, Vineet and Bavarian, Mohammad and Chen, Mark and Jun, Heewoo and Kaiser, Lukasz and Plappert, Matthias and Tworek, Jerry and Hilton, Jacob and Nakano, Reiichiro and Hesse, Christopher and Schulman, John},
  journal = {arXiv preprint arXiv:2110.14168},
  year    = {2021}
}

@article{livecodebench,
  title   = {{LiveCodeBench}: Holistic and Contamination Free Evaluation of Large Language Models for Code},
  author  = {Jain, Naman and Han, King and Gu, Alex and Li, Wen-Ding and Yan, Fanjia and Zhang, Tianjun and Wang, Sida and Solar-Lezama, Armando and Sen, Koushik and Stoica, Ion},
  journal = {arXiv preprint arXiv:2403.07974},
  year    = {2024}
}

@article{mmlupro,
  title   = {{MMLU-Pro}: A More Robust and Challenging Multi-Task Language Understanding Benchmark},
  author  = {Wang, Yubo and Ma, Xueguang and Zhang, Ge and Ni, Yuansheng and Chandra, Abhranil and Guo, Shiguang and Ren, Weiming and Arulraj, Aaran and He, Xuan and Jiang, Ziyan and Li, Tianle and Ku, Max and Wang, Kai and Zhuang, Alex and Fan, Rongqi and Yue, Xiang and Chen, Wenhu},
  journal = {arXiv preprint arXiv:2406.01574},
  year    = {2024}
}

@article{gpqa,
  title   = {{GPQA}: A Graduate-Level Google-Proof Q\&A Benchmark},
  author  = {Rein, David and Hou, Betty Li and Stickland, Asa Cooper and Petty, Jackson and Pang, Richard Yuanzhe and Dirani, Julien and Michael, Julian and Bowman, Samuel R.},
  journal = {arXiv preprint arXiv:2311.12022},
  year    = {2023}
}

@article{boolq,
  title   = {{BoolQ}: Exploring the Surprising Difficulty of Natural Yes/No Questions},
  author  = {Clark, Christopher and Lee, Kenton and Chang, Ming-Wei and Kwiatkowski, Tom and Collins, Michael and Toutanova, Kristina},
  journal = {arXiv preprint arXiv:1905.10044},
  year    = {2019}
}

@article{piqa,
  title   = {{PIQA}: Reasoning about Physical Commonsense in Natural Language},
  author  = {Bisk, Yonatan and Zellers, Rowan and Le Bras, Ronan and Gao, Jianfeng and Choi, Yejin},
  journal = {arXiv preprint arXiv:1911.11641},
  year    = {2019}
}

@article{siqa,
  title   = {{SocialIQA}: Commonsense Reasoning about Social Interactions},
  author  = {Sap, Maarten and Rashkin, Hannah and Chen, Derek and LeBras, Ronan and Choi, Yejin},
  journal = {arXiv preprint arXiv:1904.09728},
  year    = {2019}
}

@article{mmlu,
  title   = {Measuring Massive Multitask Language Understanding},
  author  = {Hendrycks, Dan and Burns, Collin and Basart, Steven and Zou, Andy and Mazeika, Mantas and Song, Dawn and Steinhardt, Jacob},
  journal = {arXiv preprint arXiv:2009.03300},
  year    = {2020}
}

@article{hellaswag,
  title   = {{HellaSwag}: Can a Machine Really Finish Your Sentence?},
  author  = {Zellers, Rowan and Holtzman, Ari and Bisk, Yonatan and Farhadi, Ali and Choi, Yejin},
  journal = {arXiv preprint arXiv:1905.07830},
  year    = {2019}
}

@article{mmmu,
  title   = {{MMMU}: A Massive Multi-discipline Multimodal Understanding and Reasoning Benchmark for Expert {AGI}},
  author  = {Yue, Xiang and Ni, Yuansheng and Zhang, Kai and Zheng, Tianyu and Liu, Ruoqi and Zhang, Ge and Stevens, Samuel and Jiang, Dongfu and Ren, Weiming and Sun, Yuxuan and Wei, Cong and Yu, Botao and Yuan, Ruibin and Sun, Renliang and Yin, Ming and Zheng, Boyuan and Yang, Zhenzhu and Liu, Yibo and Huang, Wenhao and Sun, Huan and Su, Yu and Chen, Wenhu},
  journal = {arXiv preprint arXiv:2311.16502},
  year    = {2023}
}

@article{mmbench,
  title   = {{MMBench}: Is Your Multi-modal Model an All-around Player?},
  author  = {Liu, Yuan and Duan, Haodong and Zhang, Yuanhan and Li, Bo and Zhang, Songyang and Zhao, Wangbo and Yuan, Yike and Wang, Jiaqi and He, Conghui and Liu, Ziwei and Chen, Kai and Lin, Dahua},
  journal = {arXiv preprint arXiv:2307.06281},
  year    = {2023}
}

@article{mmstar,
  title   = {Are We on the Right Way for Evaluating Large Vision-Language Models?},
  author  = {Chen, Lin and Li, Jinsong and Dong, Xiaoyi and Zhang, Pan and Zang, Yuhang and Chen, Zehui and Duan, Haodong and Wang, Jiaqi and Qiao, Yu and Lin, Dahua and Zhao, Feng},
  journal = {arXiv preprint arXiv:2403.20330},
  year    = {2024}
}

@article{chartqa,
  title   = {{ChartQA}: A Benchmark for Question Answering about Charts with Visual and Logical Reasoning},
  author  = {Masry, Ahmed and Long, Do Xuan and Tan, Jia Qing and Joty, Shafiq and Hoque, Enamul},
  journal = {arXiv preprint arXiv:2203.10244},
  year    = {2022}
}

@article{infovqa,
  title   = {{InfographicVQA}},
  author  = {Mathew, Minesh and Bagal, Viraj and Pérez Tito, Rubèn and Karatzas, Dimosthenis and Valveny, Ernest and Jawahar, C. V},
  journal = {arXiv preprint arXiv:2104.12756},
  year    = {2021}
}

@article{textvqa,
  title   = {Towards {VQA} Models That Can Read},
  author  = {Singh, Amanpreet and Natarajan, Vivek and Shah, Meet and Jiang, Yu and Chen, Xinlei and Batra, Dhruv and Parikh, Devi and Rohrbach, Marcus},
  journal = {arXiv preprint arXiv:1904.08920},
  year    = {2019}
}

@article{gqa,
  title   = {{GQA}: A New Dataset for Real-World Visual Reasoning and Compositional Question Answering},
  author  = {Hudson, Drew A. and Manning, Christopher D.},
  journal = {arXiv preprint arXiv:1902.09506},
  year    = {2019}
}

@article{vizwiz,
  title   = {{VizWiz} Grand Challenge: Answering Visual Questions from Blind People},
  author  = {Gurari, Danna and Li, Qing and Stangl, Abigale J. and Guo, Anhong and Lin, Chi and Grauman, Kristen and Luo, Jiebo and Bigham, Jeffrey P.},
  journal = {arXiv preprint arXiv:1802.08218},
  year    = {2018}
}

@article{mmerealworld,
  title   = {{MME-RealWorld}: Could Your Multimodal {LLM} Challenge High-Resolution Real-World Scenarios that are Difficult for Humans?},
  author  = {Zhang, Yi-Fan and Zhang, Huanyu and Tian, Haochen and Fu, Chaoyou and Zhang, Shuangqing and Wu, Junfei and Li, Feng and Wang, Kun and Wen, Qingsong and Zhang, Zhang and Wang, Liang and Jin, Rong and Tan, Tieniu},
  journal = {arXiv preprint arXiv:2408.13257},
  year    = {2024}
}

@article{lmmseval,
  title   = {{LMMs-Eval}: Reality Check on the Evaluation of Large Multimodal Models},
  author  = {Zhang, Kaichen and Li, Bo and Zhang, Peiyuan and Pu, Fanyi and Cahyono, Joshua Adrian and Hu, Kairui and Liu, Shuai and Zhang, Yuanhan and Yang, Jingkang and Li, Chunyuan and Liu, Ziwei},
  journal = {arXiv preprint arXiv:2407.12772},
  year    = {2024}
}

@article{vllm,
  title   = {Efficient Memory Management for Large Language Model Serving with {PagedAttention}},
  author  = {Kwon, Woosuk and Li, Zhuohan and Zhuang, Siyuan and Sheng, Ying and Zheng, Lianmin and Yu, Cody Hao and Gonzalez, Joseph E. and Zhang, Hao and Stoica, Ion},
  journal = {arXiv preprint arXiv:2309.06180},
  year    = {2023}
}

@misc{aime24,
  title        = {{AIME} 2024},
  author       = {{Hugging Face H4}},
  year         = {2024},
  howpublished = {\url{https://huggingface.co/datasets/HuggingFaceH4/aime_2024}}
}

@misc{aime25,
  title        = {{AIME} 2025},
  author       = {{Math-AI}},
  year         = {2025},
  howpublished = {\url{https://huggingface.co/datasets/math-ai/aime25}}
}

@misc{lighteval,
  title        = {{LightEval}: A lightweight framework for {LLM} evaluation},
  author       = {Habib, Nathan and Fourrier, Cl{\'e}mentine and Kydl{\'i}{\v{c}}ek, Hynek and
                  Wolf, Thomas and Tunstall, Lewis},
  year         = {2023},
  howpublished = {\url{https://github.com/huggingface/lighteval}}
}

@misc{lmevalharness,
  title     = {The Language Model Evaluation Harness},
  author    = {Gao, Leo and Tow, Jonathan and Abbasi, Baber and Biderman, Stella and Black, Sid and
               DiPofi, Anthony and Foster, Charles and Golding, Laurence and Hsu, Jeffrey and
               Le Noac'h, Alain and Li, Haonan and McDonell, Kyle and Muennighoff, Niklas and
               Ociepa, Chris and Phang, Jason and Reynolds, Laria and Schoelkopf, Hailey and
               Skowron, Aviya and Sutawika, Lintang and Tang, Eric and Thite, Anish and
               Wang, Ben and Wang, Kevin and Zou, Andy},
  publisher    = {Zenodo},
  month        = jul,
  year         = {2024},
  howpublished = {\url{https://doi.org/10.5281/zenodo.12608602}}
}

@misc{linglite2507,
  title        = {{Ling-lite-1.5-2507}},
  author       = {{Ling Team}},
  year         = {2025},
  howpublished = {\url{https://huggingface.co/inclusionAI/Ling-lite-1.5-2507}}
}
\bibliographystyle{iclr2027_conference}

\appendix

\section{Implementation}
\label{app:setup}

\S\ref{sec:dynamic} compares four published dynamic expert pruning rules. We reimplemented all four on
vLLM \citep{vllm} and extended them to the fine-grained routers of Table~\ref{tab:models}, and this appendix documents
that work: \S\ref{app:setup:patch} where the rules are applied, \S\ref{app:setup:rules} what each of them
computes, \S\ref{app:setup:calib} the offline statistics two of them need, and \S\ref{app:solve} how a
knob is turned into the target budget that makes the comparison of \S\ref{sec:dynamic} an iso-budget one.

\subsection{Where the rules are applied}
\label{app:setup:patch}

All four rules are implemented against vLLM. This is a requirement rather than a preference: over half of
our suite is long-form generation, and much of it at budgets where a pruned model rambles towards the
token limit, so the sweeps reported here are only affordable on a serving stack with paged attention and
fused expert kernels. It also means the rules are measured in the setting they would actually be deployed
in, which is what makes the throughput discussion of \S\ref{sec:speedup} meaningful.

The intervention itself is small and sits in one place, at the point where the router's scores have been
computed and the top $\ktop$ experts are about to be handed to the expert kernels. Each rule receives
that slice, in descending weight order, and returns how far down it to read; the retained gate weights
are then renormalised exactly as in Equation~\ref{eq:pruned}. Because the hook is at the selection step
rather than inside any model definition, one implementation covers all four architectures. Shared
experts are outside it by construction: every token passes through them (Equation~\ref{eq:moe}), no rule
can drop them, and they are excluded from the budgets we report, so a $\kbar$ in this paper is always a
count of routed experts. Ling-lite-1.5 has two of them, so a $\kbar$ of $2$ there costs four expert
evaluations and is not comparable with a $\kbar$ of $2$ on a model without shared experts. The code is
released.

\subsection{The four rules as implemented}
\label{app:setup:rules}

\begin{table}[t]
\centering
\small
\caption{The rules compared in \S\ref{sec:dynamic}, all re-implemented in the one router patch of
\S\ref{app:setup:patch}. Every rule keeps all $\nexp$ experts resident and changes only how many of the
$\ktop$ selected ones are evaluated per token. ``Layer budget'' is whether the rule can spend differently
at different depths; ``calib.'' whether it needs the offline statistics of \S\ref{app:setup:calib}; the
last two columns give the floor on kept experts and the single knob the budget solver of
\S\ref{app:solve} moves.}
\label{tab:methods}
\begin{tabular}{llcccl}
\toprule
Rule & Retention decision conditioned on & \makecell{layer\\budget} & \makecell{needs\\calib.}
& $k_{\min}$ & knob \\
\midrule
\fixedk{} (baseline) & a constant $k$ for all tokens               & \no  & \no  & --- & $k$ \\
\naee{}   & gate weight relative to the top-1 expert              & \no  & \no  & 2   & $\beta$ \\
\dynr{}   & cumulative retained probability mass                  & \no  & \no  & --- & $p$ \\
\diep{}   & top-1 ratio scaled by expert similarity               & \no  & \yes & 2   & $\beta$ \\
\ban{}    & layer sensitivity $+$ token sensitivity               & \yes & \yes & 3   & $\lambda$ \\
\bottomrule
\end{tabular}
\end{table}

Throughout, $w_1\ge w_2\ge\dots\ge w_{\ktop}$ are the softmax gate weights of the natively selected
experts $e_1,\dots,e_{\ktop}$ at layer $\ell$ for one token, and $k_{\min}$ is a floor on the number of
experts kept. The floor is not shared across rules: it is part of the published form of some of them and
absent from others, and we kept each rule's own convention rather than imposing one, since changing it
changes the rule. \naee{} and \diep{} use $k_{\min}{=}2$, \ban{} uses $k_{\min}{=}3$, and \dynr{} has no
floor.

\paragraph{\naee{}.} A flat threshold relative to the leading expert, applied as a prefix. Rank $i$ is
kept if it is below the floor, or if no rank from the floor up to and including $i$ has fallen below the
threshold:
\begin{equation}
\text{keep } i \iff i < k_{\min} \quad\text{or}\quad w_j \ge \beta\,w_1
\;\;\text{for all } j \text{ with } k_{\min} \le j \le i .
\label{eq:naee}
\end{equation}
Equivalently, reading down the ranking, retention stops at the first expert whose weight is less than a
$\beta$ fraction of the top-1 weight. The rule has one knob, $\beta$, and needs no calibration.

\paragraph{\dynr{}.} A threshold on cumulative retained mass rather than on any single pair of experts.
The kept set is the shortest prefix of the ranking whose weights reach a target $p$:
\begin{equation}
k(\vx,\ell) \;=\; \min\Big\{\, i \;:\; \textstyle\sum_{j\le i} \hat{w}_j \ge p \,\Big\},
\label{eq:dynroute}
\end{equation}
falling back to $\ktop$ if the target is never reached. Two readings of $\hat{w}$ are implemented: the
softmax weights as they are, which is the published form, and the same weights renormalised over the
$\ktop$ retained ranks first, which makes $p$ a fraction of the mass the router actually routes rather
than of the whole distribution. The configurations reported in Table~\ref{tab:iso} use the renormalised
reading, since it is the one whose $p$ keeps its meaning across models with different $\ktop$. The rule
has one knob and needs no calibration.

\paragraph{\diep{}.} \naee{}'s threshold, scaled per expert by how similar that expert is to the leading
one. With $\mathrm{sim}_\ell(\cdot,\cdot)$ a calibrated similarity between expert outputs at layer
$\ell$ and $\overline{\mathrm{sim}}_\ell$ its mean over expert pairs at that layer,
\begin{equation}
\gamma_i \;=\; \Big(\frac{\mathrm{sim}_\ell(e_1,e_i)}{\overline{\mathrm{sim}}_\ell}\Big)^{\!\alpha},
\qquad \gamma_1 \equiv 1,
\qquad
\text{keep } i \iff i < k_{\min} \;\text{ or }\; w_i \ge \beta\,\gamma_i\,w_1 .
\label{eq:diep}
\end{equation}
An expert that duplicates the leading one faces a higher bar and is dropped sooner; one that
contributes something different is kept on a smaller weight. The exponent $\alpha$ is ours, not the
published rule's, and interpolates: $\alpha{=}1$ is the rule as published, $\alpha{=}0$ removes the
similarity term and reduces Equation~\ref{eq:diep} exactly to Equation~\ref{eq:naee}, and intermediate
values damp the spread of $\gamma$ without removing it. Rows marked \diep{}-d in
Appendix~\ref{app:full:iso} use $\alpha<1$. Whether the threshold is applied per rank independently or as
a prefix is also a choice; Appendix~\ref{app:diep} explains why, and reports both.

One limitation of our \diep{} rows has to be stated plainly, because it changes how they should be read.
The published method pairs this skipping criterion with a differentiable pruning objective; we implement
only the criterion. Equation~\ref{eq:diep} is therefore our generalisation of one component of
\diep{} to a wide router, and the \diep{} rows in Appendix~\ref{app:full:iso} are evidence about that
component under that generalisation, not about the quality \diep{} as published can reach. This is
also why we do not present those rows as a ranking of the method against the others in the main text.

\paragraph{\ban{}.} The only rule here that varies the budget with depth as well as with the token. It
converts two sensitivity signals into an integer expert count. The token signal is how much of the
routed mass the leading three experts hold,
\begin{equation}
r \;=\; \frac{\sum_{i\le3} w_i}{\sum_{i\le \ktop} w_i},
\qquad
s_{\mathrm{tok}} \;=\; \mathrm{clamp}\Big(\frac{r_{\max}-r}{r_{\max}-r_{\min}},\,0,\,1\Big),
\label{eq:ban_tok}
\end{equation}
so a token whose distribution is peaked scores low and a flat one scores high. This is combined with a
calibrated per-layer sensitivity $s_\ell$ and mapped onto the available range of expert counts:
\begin{equation}
s \;=\; \mathrm{clamp}\Big(\tfrac{\lambda}{2}\big(s_\ell + s_{\mathrm{tok}}\big),\,0,\,1\Big),
\quad
k(\vx,\ell) \;=\; \mathrm{clamp}\Big(\mathrm{round}\big(k_{\min} + (\ktop-k_{\min})\,s\big),\,
k_{\min},\,\ktop\Big).
\label{eq:ban}
\end{equation}
The single knob $\lambda$ scales the combined sensitivity and therefore the average budget. Note the
direction of Equation~\ref{eq:ban_tok}: it is the tokens whose routers are least decided that receive
the most experts, which is the mechanism \S\ref{sec:dynamic:results} appeals to, and $r_{\min}$, $r_{\max}$
and $s_\ell$ all come from calibration.

\subsection{Offline calibration}
\label{app:setup:calib}

\diep{} and \ban{} need statistics that cannot be read off a single token's router output. Both are
collected once per checkpoint, before any evaluation, on $512$ documents of C4 under HuggingFace
Transformers, and stored as an artifact that the serving-time patch loads. No evaluation data is used
for calibration, and no evaluation run searches over hyperparameters.

For \diep{} the artifact holds, per MoE layer, the pairwise linear CKA between expert outputs on the
calibration tokens, which is the $\mathrm{sim}_\ell$ of Equation~\ref{eq:diep}, together with its mean
over pairs. It also holds a per-layer median of $w_2/w_1$, which the published rule uses as a second
scaling factor; we fold that level into $\beta$ instead, since $\beta$ is what the budget solver of
\S\ref{app:solve} moves, and report the variant that keeps it separately in
Appendix~\ref{app:diep}.

For \ban{} the artifact holds the per-layer sensitivity $s_\ell$ and the range $[r_{\min}, r_{\max}]$.
The former is obtained by comparing, at each layer in turn, the layer output under full top-$\ktop$
routing against its output when only the leading three experts are kept, measuring the divergence
between the resulting next-token distributions, and normalising across layers; the latter is the
observed minimum and maximum of $r$ over the calibration tokens.

\subsection{Solving each knob to a target budget}
\label{app:solve}

Every rule is characterised by the budget it spends rather than by its knob
(\S\ref{sec:dynamic:rules}), so each knob has to be set to hit a target $\kbar$ before the run, not
after. Doing this by trial evaluation would be prohibitive: a single configuration of the largest model
is a multi-hour run. We therefore solve each knob offline against measured routing statistics.

For \naee{} and \dynr{} the map from knob to average expert count is estimated directly, by replaying
the retention criterion over router outputs recorded on the calibration corpus across a grid of knob
values, and the value whose predicted average is closest to the target is used. For \diep{} the same
grid is not available, because the threshold depends on $\gamma$; instead the weight distribution is
summarised by fitting a Beta survival function to the \naee{} sweep, the average under
Equation~\ref{eq:diep} is predicted by integrating that survival function against the calibrated
$\gamma$ distribution, and $\beta$ is found by binary search. For \ban{} the token sensitivity of
Equation~\ref{eq:ban_tok} is fitted with a Beta distribution and the rounding in
Equation~\ref{eq:ban} is inverted numerically for $\lambda$.

Because a threshold is not a budget (\S\ref{sec:dynamic:rules}), every knob is solved twice, once
against the generative prompt distribution and once against the likelihood-scored one. This is why a
rule in Table~\ref{tab:iso} and in Appendix~\ref{app:full:iso} carries two measured budgets rather than
one. The residual between the solved target and the realised $\kbar$ is what the budget columns of
those tables report, and it is the reason rules that overspend their tier by more than $10\%$ are
excluded from the main-text comparison rather than quietly compared at a cheaper budget.

\section{Evaluation protocol: how a sweep can silently be wrong}
\label{app:protocol}

This appendix records the measurement problems we hit, because each produced results that looked
entirely plausible and each invalidates a comparison in a different way.

\paragraph{Comparing at equal hyperparameter rather than equal budget.}
A threshold is not a budget. The same setting yields a different average expert count depending on the
routing distributions it acts on, and those distributions differ between long-form generation and
likelihood-scored QA. Two rules run at the same nominal coefficient can therefore be spending different
amounts of computation, so a comparison between them reflects the gap in budget as much as the
difference in allocation. We solve each rule's threshold separately in the two regimes and report the
measured $\kbar$ for every configuration, so that the comparison in \S\ref{sec:dynamic} is between rules
at a matched budget rather than at a matched hyperparameter.

\paragraph{Reference rows that are noisy in the favourable direction.} Every margin in
\S\ref{sec:dynamic} is measured against a \fixedk{} row, so a \fixedk{} row that reads high makes every
rule look worse and one that reads low makes every rule look better. Some \fixedk{} configurations were
measured twice over the course of the study, and the two readings disagree by up to a point. Our
convention is to keep the lower of the two, so that no rule can be credited with beating a reference
that happened to read high. On Qwen3-30B-A3B the pairs are $74.79$ against $74.09$ at $k{=}6$, $73.67$
against $72.81$ at $k{=}5$, $57.65$ against $56.44$ at $k{=}3$ and $15.80$ against $15.74$ at $k{=}2$;
on Ling-lite-1.5 they are $59.39$ against $58.82$ at $k{=}6$, $58.90$ against $58.38$ at $k{=}5$,
$57.06$ against $57.02$ at $k{=}4$, $52.60$ against $51.49$ at $k{=}3$ and $34.50$ against $33.47$ at
$k{=}2$. In every case the newer reading was the higher one, and in every case we kept the older and
lower one. The margins reported in Table~\ref{tab:iso} are therefore conservative in the direction
unfavourable to our own conclusion, which is the direction we want them to err in given that the
conclusion is that \fixedk{} is hard to beat.

\paragraph{Rules that undershoot the budget rather than overshoot it.} The exclusion described in
\S\ref{app:solve} catches configurations that spend more than their tier. The opposite case also
occurs and is not symmetric: a rule spending less than \fixedk{} at the same tier, by up to $0.95$
experts in the worst case we have, can score below \fixedk{} for the trivial reason that it was given
less. Ten rows on Qwen3-30B-A3B and two on GPT-OSS-20B are in that position. They are left in
Appendix~\ref{app:full:iso} with their measured budgets visible rather than dropped, since dropping
them would hide configurations that a reader may want to see, but a deficit on such a row is not
evidence about the rule.

\section{Generalising a top-2 skipping rule to top-$\ktop$ routing}
\label{app:diep}

\diep{} was published for a router that selects two experts per token, and the shape of the rule
reflects that. With $\ktop{=}2$ there is exactly one decision to make, whether to keep $e_2$, and
exactly one similarity to consult, that between $e_1$ and $e_2$. Both the threshold and the calibrated
statistic are therefore defined on the rank-2 expert specifically. Applying the rule to a router that
selects eight or ten experts requires deciding what the rank-$i$ analogue is, and the choice is not
forced by the original.

Our generalisation is the one in Equation~\ref{eq:diep}: every rank $i>1$ is compared against the
leading expert with its own similarity factor $\gamma_i$, and each rank is judged independently. The
alternative we also implemented applies the same per-rank threshold but stops at the first failure, as
\naee{} does, which keeps the retained set a prefix of the ranking. The two differ whenever a
low-similarity expert sits below a high-similarity one: the independent form can keep rank $5$ after
dropping rank $3$, the prefix form cannot. We report the independent form in the main text because it is
the more faithful reading of a rule whose whole content is a per-expert comparison, and note that
neither choice rescues the rule at the budgets of Table~\ref{tab:iso}.

The rule's own calibrated level is also rank-2-specific. The published form multiplies the threshold by
a per-layer median of $w_2/w_1$, a quantity with no rank-$i$ analogue that is not simply the
distribution of $w_i/w_1$, which changes shape with $\ktop$. We therefore absorb the level into $\beta$
and let the budget solver of \S\ref{app:solve} set it, which is also what makes the comparison at
matched $\kbar$ possible at all. Keeping the factor explicit and setting $\beta{=}1$ would stay closer
to the original expression, but it pins the operating point to whatever level the calibration statistic
implies and leaves no free parameter to solve, so such a configuration could not be placed at a matched
budget alongside the other rules.

Two observations from the results, since they bear on the reading of Appendix~\ref{app:full:iso} rather
than on our implementation choices. The similarity term's contribution is not small: on Qwen3-30B-A3B at
the conservative budget, damping it with $\alpha<1$ moves the suite mean from $54.32$ to $61.60$, and
removing it entirely, which is \naee{}, gives $63.92$. And the size of the deficit does not track router
width in the way one might expect. Against \fixedk{} at the conservative budget it is $-11.27$ on
GPT-OSS-20B at $\ktop{=}4$ and $-11.11$ on Qwen3-30B-A3B at $\ktop{=}8$, but only $-2.38$ on
Ling-lite-1.5 at $\ktop{=}6$ and $-2.16$ on Qwen3-Next-80B at $\ktop{=}10$, the widest router in the
set. Whatever our generalisation costs, it is not a simple function of how many experts there are to
generalise over, and we do not have a mechanism for the pattern.

\section{Full results: uniform truncation}
\label{app:full:fixedk}

Tables~\ref{tab:app:fixedk} and~\ref{tab:app:fixedk:add} give every \fixedk{} budget on every
dataset for all nine models, which is the data behind Figure~\ref{fig:redundancy} and behind the task-family curves discussed in
\S\ref{sec:redundancy}. Reading down a block shows the two features the main text draws on: the near-flat
region at the top, and the fact that the collapse arrives at a fraction of the native budget rather than
at a fixed number of experts, so that $k{=}2$ is catastrophic on the $\ktop{=}8$ model and almost free on
the $\ktop{=}4$ one.

\begin{table}[t]
\centering
\scriptsize
\setlength{\tabcolsep}{0.92pt}
\caption{Uniform truncation, every budget and every dataset, on the four core models. Each row is one value of $k$ applied to all tokens and all layers; the top row of each block is the unpruned model. Scores are each dataset's primary metric as a percentage and the last column is the unweighted mean over the datasets that were run: a dash marks a cell that was not, and it is left out of that row's mean. Green rows are above half the native $\ktop$, where uniform truncation needs no rule to help it; orange rows are at or below it, past the crossover of \S\ref{sec:dynamic}. This table and Table~\ref{tab:app:fixedk:add} are what Figure~\ref{fig:redundancy} is drawn from.}
\label{tab:app:fixedk}
\begin{tabularx}{\linewidth}{L cccc ccccc cc c}
\toprule
& \multicolumn{4}{c}{Knowledge QA}& \multicolumn{5}{c}{Math \& code reasoning}& \multicolumn{2}{c}{Knowledge \& general} & \\
\cmidrule(lr){2-5}\cmidrule(lr){6-10}\cmidrule(lr){11-12}
& ARC-c & ARC-e & WG & OBQA& MATH & AIME24 & AIME25 & GSM8K & LCB& MMLU-P & GPQA-D & Avg \\
\midrule
\rowcolor{modelfill}\multicolumn{13}{l}{\textbf{Qwen3-30B-A3B-Instruct}, natively $\ktop{=}8$}\\
\rowcolor{consfill}\quad unpruned, $k{=}8$ & 60.8 & 85.0 & 73.2 & 32.0 & 89.6 & 74.1 & 61.4 & 94.4 & 41.1 & 74.2 & 56.6 & 67.49 \\
\rowcolor{consfill}\quad \fixedk{}, $k{=}7$ & 60.3 & 84.0 & 73.2 & 33.2 & 89.2 & 74.6 & 61.3 & 94.7 & 41.3 & 74.5 & 56.6 & 67.54 \\
\rowcolor{consfill}\quad \fixedk{}, $k{=}6$ & 59.2 & 82.8 & 71.3 & 31.4 & 90.4 & 72.8 & 59.3 & 95.0 & 44.6 & 74.4 & 53.5 & 66.79 \\
\rowcolor{consfill}\quad \fixedk{}, $k{=}5$ & 54.9 & 81.4 & 69.5 & 31.0 & 91.0 & 72.1 & 55.4 & 93.9 & 41.7 & 73.8 & 55.0 & 65.43 \\
\rowcolor{aggrfill}\quad \fixedk{}, $k{=}4$ & 51.3 & 77.2 & 65.1 & 31.2 & 87.8 & 62.5 & 47.9 & 93.3 & 40.6 & 72.3 & 54.0 & 62.11 \\
\rowcolor{aggrfill}\quad \fixedk{}, $k{=}3$ & 44.6 & 71.8 & 58.0 & 27.0 & 80.2 & 40.0 & 31.1 & 88.5 & 25.7 & 67.0 & 48.5 & 52.95 \\
\rowcolor{aggrfill}\quad \fixedk{}, $k{=}2$ & 28.7 & 51.0 & 50.6 & 21.4 & 24.6 & 0.0 & 0.0 & 22.9 & 1.1 & 22.1 & 27.3 & 22.70 \\
\rowcolor{aggrfill}\quad \fixedk{}, $k{=}1$ & 22.2 & 26.4 & 48.9 & 14.0 & 2.0 & 0.0 & 0.2 & 1.0 & 0.0 & 7.4 & 28.8 & 13.72 \\
\midrule
\rowcolor{modelfill}\multicolumn{13}{l}{\textbf{Qwen3-Next-80B-A3B-Instruct}, natively $\ktop{=}10$}\\
\rowcolor{consfill}\quad unpruned, $k{=}10$ & 63.1 & 86.9 & 76.2 & 34.2 & 88.4 & 79.7 & 67.1 & 95.2 & 53.1 & 82.1 & 74.2 & 72.75 \\
\rowcolor{consfill}\quad \fixedk{}, $k{=}9$ & 63.4 & 87.0 & 75.7 & 34.2 & 89.4 & 79.4 & 68.5 & 95.6 & 53.1 & 82.2 & 76.3 & 73.16 \\
\rowcolor{consfill}\quad \fixedk{}, $k{=}8$ & 63.1 & 86.5 & 74.5 & 34.6 & 90.0 & 81.8 & 68.5 & 95.8 & 54.9 & 82.4 & 68.7 & 72.80 \\
\rowcolor{consfill}\quad \fixedk{}, $k{=}7$ & 63.1 & 86.0 & 74.6 & 35.8 & 90.4 & 81.8 & 68.8 & 95.2 & 54.9 & 82.0 & 74.2 & 73.35 \\
\rowcolor{consfill}\quad \fixedk{}, $k{=}6$ & 63.1 & 85.6 & 73.9 & 33.6 & 90.8 & 82.2 & 66.6 & 95.7 & 54.9 & 82.1 & 76.3 & 73.16 \\
\rowcolor{aggrfill}\quad \fixedk{}, $k{=}5$ & 59.6 & 84.9 & 71.0 & 32.6 & 90.8 & 81.1 & 66.0 & 94.9 & 54.9 & 81.8 & 69.2 & 71.53 \\
\rowcolor{aggrfill}\quad \fixedk{}, $k{=}4$ & 55.9 & 82.2 & 67.2 & 30.2 & 88.0 & 74.8 & 58.8 & 93.1 & 49.1 & 80.7 & 72.7 & 68.43 \\
\rowcolor{aggrfill}\quad \fixedk{}, $k{=}3$ & 50.0 & 77.4 & 62.1 & 29.8 & 87.6 & 57.8 & 40.8 & 86.1 & 34.9 & 78.0 & 66.7 & 61.02 \\
\rowcolor{aggrfill}\quad \fixedk{}, $k{=}2$ & 36.8 & 61.1 & 56.1 & 22.2 & 69.0 & 12.0 & 7.7 & 75.0 & 7.4 & 62.3 & 44.4 & 41.27 \\
\rowcolor{aggrfill}\quad \fixedk{}, $k{=}1$ & 22.4 & 35.0 & 49.6 & 16.4 & 1.6 & 0.0 & 0.0 & 0.9 & 0.0 & 8.5 & 28.3 & 14.79 \\
\midrule
\rowcolor{modelfill}\multicolumn{13}{l}{\textbf{Ling-lite-1.5}, natively $\ktop{=}6$}\\
\rowcolor{consfill}\quad unpruned, $k{=}6$ & 60.2 & 82.3 & 70.3 & 31.6 & 89.2 & 40.9 & 28.1 & 91.4 & 34.3 & 69.4 & 56.6 & 59.48 \\
\rowcolor{consfill}\quad \fixedk{}, $k{=}5$ & 59.9 & 81.9 & 69.4 & 31.0 & 91.0 & 43.0 & 27.7 & 90.3 & 33.7 & 69.1 & 58.1 & 59.55 \\
\rowcolor{consfill}\quad \fixedk{}, $k{=}4$ & 58.3 & 81.2 & 66.1 & 30.8 & 89.0 & 38.7 & 26.0 & 90.5 & 33.7 & 67.4 & 55.6 & 57.94 \\
\rowcolor{aggrfill}\quad \fixedk{}, $k{=}3$ & 52.3 & 79.4 & 63.9 & 29.4 & 86.0 & 31.5 & 24.2 & 87.5 & 28.6 & 63.3 & 46.5 & 53.87 \\
\rowcolor{aggrfill}\quad \fixedk{}, $k{=}2$ & 38.9 & 69.6 & 57.4 & 23.0 & 70.4 & 5.0 & 8.4 & 72.4 & 8.0 & 42.3 & 38.4 & 39.44 \\
\rowcolor{aggrfill}\quad \fixedk{}, $k{=}1$ & 21.0 & 29.0 & 48.5 & 15.2 & 1.2 & 0.0 & 0.2 & 1.0 & 0.0 & 7.5 & 27.8 & 13.76 \\
\midrule
\rowcolor{modelfill}\multicolumn{13}{l}{\textbf{GPT-OSS-20B}, natively $\ktop{=}4$}\\
\rowcolor{consfill}\quad unpruned, $k{=}4$ & 45.3 & 77.5 & 66.9 & 27.2 & 90.6 & 73.7 & 71.7 & 86.2 & 58.3 & 74.3 & 64.6 & 66.94 \\
\rowcolor{consfill}\quad \fixedk{}, $k{=}3$ & 44.8 & 76.7 & 65.4 & 27.4 & 91.0 & 77.4 & 74.5 & 87.0 & 61.7 & 73.6 & 65.7 & 67.74 \\
\rowcolor{aggrfill}\quad \fixedk{}, $k{=}2$ & 44.3 & 75.9 & 64.1 & 26.0 & 86.6 & 73.9 & 69.4 & 88.7 & 48.0 & 70.5 & 65.2 & 64.79 \\
\rowcolor{aggrfill}\quad \fixedk{}, $k{=}1$ & 23.8 & 45.0 & 50.2 & 18.0 & 13.2 & 0.4 & 0.6 & 7.6 & 1.1 & 11.9 & 25.3 & 17.92 \\
\bottomrule
\end{tabularx}
\end{table}

\begin{table}[t]
\centering
\scriptsize
\setlength{\tabcolsep}{0.92pt}
\caption{Uniform truncation, every budget and every dataset, on the five further architectures. Each row is one value of $k$ applied to all tokens and all layers; the top row of each block is the unpruned model. Scores are each dataset's primary metric as a percentage and the last column is the unweighted mean over the datasets that were run: a dash marks a cell that was not, and it is left out of that row's mean. Green rows are above half the native $\ktop$, where uniform truncation needs no rule to help it; orange rows are at or below it, past the crossover of \S\ref{sec:dynamic}. Columns, tints and the reference row follow Table~\ref{tab:app:fixedk}.}
\label{tab:app:fixedk:add}
\begin{tabularx}{\linewidth}{L cccc ccccc cc c}
\toprule
& \multicolumn{4}{c}{Knowledge QA}& \multicolumn{5}{c}{Math \& code reasoning}& \multicolumn{2}{c}{Knowledge \& general} & \\
\cmidrule(lr){2-5}\cmidrule(lr){6-10}\cmidrule(lr){11-12}
& ARC-c & ARC-e & WG & OBQA& MATH & AIME24 & AIME25 & GSM8K & LCB& MMLU-P & GPQA-D & Avg \\
\midrule
\rowcolor{modelfill}\multicolumn{13}{l}{\textbf{MiniMax-M2.7}, natively $\ktop{=}8$}\\
\rowcolor{consfill}\quad unpruned, $k{=}8$ & 70.1 & 86.8 & 76.2 & 38.8 & 90.2 & 81.5 & 81.7 & 91.6 & 47.3 & 82.0 & 85.4 & 75.60 \\
\rowcolor{consfill}\quad \fixedk{}, $k{=}7$ & 69.7 & 86.3 & 74.6 & 37.8 & 89.6 & 80.8 & 82.1 & 92.0 & 44.0 & 81.9 & 87.9 & 75.15 \\
\rowcolor{consfill}\quad \fixedk{}, $k{=}6$ & 69.2 & 85.6 & 74.4 & 37.4 & 88.8 & 78.3 & 80.6 & 91.5 & 42.3 & 81.5 & 84.3 & 73.99 \\
\rowcolor{consfill}\quad \fixedk{}, $k{=}5$ & 66.3 & 84.3 & 70.0 & 37.0 & 89.0 & 73.2 & 75.8 & 91.7 & 44.6 & 80.9 & 84.3 & 72.46 \\
\rowcolor{aggrfill}\quad \fixedk{}, $k{=}4$ & 63.1 & 82.9 & 68.0 & 35.8 & 85.4 & 54.2 & 42.9 & 90.1 & 33.1 & 78.7 & 78.8 & 64.82 \\
\rowcolor{aggrfill}\quad \fixedk{}, $k{=}3$ & 52.3 & 75.9 & 60.5 & 32.6 & 65.8 & 4.2 & 10.8 & 83.6 & 9.1 & 59.4 & 51.0 & 45.93 \\
\rowcolor{aggrfill}\quad \fixedk{}, $k{=}2$ & 34.9 & 53.8 & 53.2 & 22.2 & 3.0 & 0.0 & 0.0 & 2.3 & 0.0 & 9.4 & 24.2 & 18.45 \\
\rowcolor{aggrfill}\quad \fixedk{}, $k{=}1$ & 20.6 & 26.1 & 49.6 & 15.6 & 1.4 & 0.0 & 0.0 & 2.0 & 0.0 & 5.2 & 23.2 & 13.06 \\
\midrule
\rowcolor{modelfill}\multicolumn{13}{l}{\textbf{DeepSeek-V2-Lite-Chat}, natively $\ktop{=}6$}\\
\rowcolor{consfill}\quad unpruned, $k{=}6$ & 50.9 & 80.1 & 71.4 & 36.2 & 25.0 & --- & --- & 63.1 & 10.3 & 27.4 & 28.3 & 43.63 \\
\rowcolor{consfill}\quad \fixedk{}, $k{=}5$ & 50.3 & 79.4 & 70.8 & 35.6 & 24.6 & --- & --- & 61.7 & 10.9 & 28.2 & 30.3 & 43.53 \\
\rowcolor{consfill}\quad \fixedk{}, $k{=}4$ & 50.3 & 79.1 & 70.0 & 35.4 & 24.0 & --- & --- & 61.7 & 10.3 & 26.9 & 30.8 & 43.17 \\
\rowcolor{aggrfill}\quad \fixedk{}, $k{=}3$ & 47.6 & 77.4 & 69.9 & 33.8 & 26.2 & --- & --- & 56.1 & 6.9 & 24.3 & 26.3 & 40.94 \\
\rowcolor{aggrfill}\quad \fixedk{}, $k{=}2$ & 43.1 & 73.4 & 67.6 & 28.8 & 16.6 & --- & --- & 47.8 & 5.7 & 18.8 & 25.8 & 36.40 \\
\rowcolor{aggrfill}\quad \fixedk{}, $k{=}1$ & 33.6 & 63.7 & 61.8 & 25.2 & 3.2 & --- & --- & 14.9 & 1.1 & 11.1 & 25.3 & 26.66 \\
\midrule
\rowcolor{modelfill}\multicolumn{13}{l}{\textbf{Hy3}, natively $\ktop{=}8$}\\
\rowcolor{consfill}\quad unpruned, $k{=}8$ & 73.4 & 88.3 & 80.1 & 50.6 & 81.4 & 93.8 & 92.1 & 95.8 & 67.4 & 86.5 & 85.3 & 81.34 \\
\rowcolor{consfill}\quad \fixedk{}, $k{=}7$ & 72.9 & 88.1 & 78.5 & 50.4 & 81.0 & 93.3 & 89.8 & 95.8 & 65.7 & 86.6 & 84.3 & 80.58 \\
\rowcolor{consfill}\quad \fixedk{}, $k{=}6$ & 71.2 & 87.6 & 77.1 & 49.0 & 81.6 & 92.5 & 87.9 & 96.0 & 64.0 & 86.6 & 82.8 & 79.68 \\
\rowcolor{consfill}\quad \fixedk{}, $k{=}5$ & 69.9 & 86.0 & 76.2 & 47.6 & 80.8 & 90.6 & 82.1 & 95.5 & 57.7 & 86.5 & 81.8 & 77.70 \\
\rowcolor{aggrfill}\quad \fixedk{}, $k{=}4$ & 66.1 & 84.8 & 73.6 & 45.2 & 80.2 & 82.5 & 67.9 & 95.3 & 48.0 & 86.1 & 79.8 & 73.60 \\
\rowcolor{aggrfill}\quad \fixedk{}, $k{=}3$ & 61.4 & 79.8 & 67.6 & 44.0 & 74.0 & 32.3 & 25.6 & 94.7 & 24.0 & 83.9 & 60.1 & 58.85 \\
\rowcolor{aggrfill}\quad \fixedk{}, $k{=}2$ & 46.5 & 66.1 & 58.1 & 34.2 & 1.4 & 0.0 & 0.0 & --- & 0.0 & --- & 24.8 & 25.67 \\
\rowcolor{aggrfill}\quad \fixedk{}, $k{=}1$ & 25.4 & 31.1 & 51.0 & 27.8 & 1.6 & 0.0 & 0.0 & 0.8 & 0.0 & 5.8 & 25.2 & 15.33 \\
\midrule
\rowcolor{modelfill}\multicolumn{13}{l}{\textbf{DeepSeek-V4-Flash-0731}, natively $\ktop{=}6$}\\
\rowcolor{consfill}\quad unpruned, $k{=}6$ & 67.3 & 87.8 & 78.8 & 47.8 & 88.8 & 92.1 & 91.7 & 75.0 & 29.7 & 86.3 & 87.4 & 75.69 \\
\rowcolor{consfill}\quad \fixedk{}, $k{=}5$ & 66.2 & 86.8 & 78.1 & 48.8 & 88.0 & 92.9 & 93.5 & 76.8 & 32.0 & 86.7 & 86.9 & 76.07 \\
\rowcolor{consfill}\quad \fixedk{}, $k{=}4$ & 62.2 & 85.2 & 75.0 & 46.0 & 85.8 & 92.1 & 92.9 & 75.7 & 28.6 & 86.5 & 89.9 & 74.53 \\
\rowcolor{aggrfill}\quad \fixedk{}, $k{=}3$ & 58.0 & 81.7 & 70.8 & 44.6 & 85.4 & 91.0 & 89.2 & 81.7 & 29.1 & 85.3 & 86.4 & 73.02 \\
\rowcolor{aggrfill}\quad \fixedk{}, $k{=}2$ & 48.8 & 70.5 & 63.1 & 37.8 & 77.8 & 74.4 & 68.5 & 88.7 & 44.0 & 75.3 & 72.7 & 65.61 \\
\rowcolor{aggrfill}\quad \fixedk{}, $k{=}1$ & 27.6 & 39.3 & 52.3 & 27.2 & 1.2 & 0.0 & 0.0 & 1.7 & 0.0 & 6.1 & 12.6 & 15.27 \\
\midrule
\rowcolor{modelfill}\multicolumn{13}{l}{\textbf{Gemma 4 26B-A4B}, natively $\ktop{=}8$}\\
\rowcolor{consfill}\quad unpruned, $k{=}8$ & 45.6 & 67.2 & 56.7 & 31.6 & 85.2 & 91.5 & 86.9 & 92.2 & 78.9 & 84.6 & 80.3 & 72.77 \\
\rowcolor{consfill}\quad \fixedk{}, $k{=}7$ & 44.3 & 65.1 & 56.2 & 31.0 & 85.0 & 92.1 & 87.5 & 92.3 & 76.6 & 84.9 & 79.3 & 72.20 \\
\rowcolor{consfill}\quad \fixedk{}, $k{=}6$ & 42.6 & 63.6 & 53.9 & 31.6 & 84.6 & 89.8 & 85.4 & 92.2 & 72.0 & 84.5 & 79.8 & 70.92 \\
\rowcolor{consfill}\quad \fixedk{}, $k{=}5$ & 39.1 & 59.9 & 53.0 & 28.8 & 84.8 & 88.8 & 84.2 & 92.5 & 73.7 & 84.0 & 77.3 & 69.64 \\
\rowcolor{aggrfill}\quad \fixedk{}, $k{=}4$ & 35.8 & 51.3 & 52.6 & 29.4 & 85.6 & 84.6 & 71.9 & 91.5 & 73.1 & 82.5 & 72.7 & 66.45 \\
\rowcolor{aggrfill}\quad \fixedk{}, $k{=}3$ & 32.8 & 45.0 & 52.2 & 31.0 & 82.6 & 51.5 & 41.2 & 79.8 & 62.9 & 77.3 & 69.7 & 56.90 \\
\rowcolor{aggrfill}\quad \fixedk{}, $k{=}2$ & 29.7 & 37.4 & 52.0 & 29.4 & 40.6 & 2.1 & 0.2 & 47.5 & 20.0 & 55.0 & 41.4 & 32.30 \\
\rowcolor{aggrfill}\quad \fixedk{}, $k{=}1$ & 25.0 & 29.3 & 50.7 & 31.2 & 3.6 & 0.0 & 0.0 & 2.5 & 0.0 & 10.7 & 30.3 & 16.66 \\
\bottomrule
\end{tabularx}
\end{table}

\section{Full results: every rule at both budgets}
\label{app:full:iso}

Tables~\ref{tab:app:iso0} to~\ref{tab:app:iso3} expand Table~\ref{tab:iso}. For each core model they
give every rule we evaluated at each of the two budget tiers, per dataset, with \fixedk{} at the head of
each tier so that the comparison the main text summarises can be read directly. Table~\ref{tab:iso}
reports the best rule per tier; these tables report all of them, which is what supports two claims made
in passing in \S\ref{sec:dynamic:results}. The first is that the ordering of rules changes between the
two tiers on the same model. The second is that the spread between rules at a matched budget is wider
than the gap between the best rule and \fixedk{}, so a comparison that reports one rule against an
unpruned baseline, without \fixedk{} at the same $\kbar$, is not measuring what it appears to measure.

Budgets are the measured $\kbar$ of the run, generative and QA solved separately
(\S\ref{app:solve}). A rule missing from a tier either has no configuration that lands within $10\%$ of
it or did not complete; those cases are not silently replaced by a cheaper configuration.

\begin{table}[t]
\centering
\scriptsize
\setlength{\tabcolsep}{0.92pt}
\caption{Every rule at both budget tiers on \textbf{Qwen3-30B-A3B} ($\ktop{=}8$), per dataset. Uniform truncation is the first row of each tier so that the two can be read together. The budget beside a rule is its measured $\kbar$, generative\,/\,QA. A dagger marks a configuration that spends more than $10\%$ over its tier: it is shown for completeness but is not an iso-budget comparison and was not eligible for Table~\ref{tab:iso}. The best eligible mean is bold where it beats \fixedk{}, which is the row Table~\ref{tab:iso} carries.}
\label{tab:app:iso0}
\begin{tabularx}{\linewidth}{L cccc ccccc cc c}
\toprule
& \multicolumn{4}{c}{Knowledge QA}& \multicolumn{5}{c}{Math \& code reasoning}& \multicolumn{2}{c}{Knowledge \& general} & \\
\cmidrule(lr){2-5}\cmidrule(lr){6-10}\cmidrule(lr){11-12}
& ARC-c & ARC-e & WG & OBQA& MATH & AIME24 & AIME25 & GSM8K & LCB& MMLU-P & GPQA-D & Avg \\
\midrule
\quad unpruned, $k{=}8$ & 60.8 & 85.0 & 73.2 & 32.0 & 89.6 & 74.1 & 61.4 & 94.4 & 41.1 & 74.2 & 56.6 & 67.49 \\
\rowcolor{consfill}\multicolumn{13}{l}{\quad\textit{Conservative budget}, $\kbar\!\approx\!5$ ($\rbudget{=}0.62$)}\\
\rowcolor{consfill}\quad \fixedk{}, $k{=}5$ & 54.9 & 81.4 & 69.5 & 31.0 & 91.0 & 72.1 & 55.4 & 93.9 & 41.7 & 73.8 & 55.0 & 65.43 \\
\rowcolor{consfill}\quad \naee{}, $5.13/5.04$ & 51.5 & 80.3 & 70.4 & 32.6 & 81.0 & 69.1 & 53.6 & 92.6 & 43.4 & 72.5 & 56.1 & 63.92 \\
\rowcolor{consfill}\quad \diep{}, $4.84/5.07$ & 52.6 & 79.1 & 67.2 & 31.2 & 81.8 & 35.3 & 26.5 & 92.7 & 14.9 & 70.2 & 46.0 & 54.32 \\
\rowcolor{consfill}\quad \diep{}-d, $4.93/5.08$ & 51.8 & 79.2 & 69.1 & 30.8 & 89.4 & 60.9 & 49.7 & 92.5 & 32.0 & 71.2 & 51.0 & 61.60 \\
\rowcolor{consfill}\quad \dynr{}, $4.91/5.09$ & 56.5 & 81.4 & 71.6 & 31.0 & 89.6 & 73.2 & 55.2 & 94.2 & 41.7 & 73.2 & 55.1 & \textbf{65.70} \\
\rowcolor{consfill}\quad \ban{}, $5.00/5.03$ & 55.3 & 81.8 & 69.7 & 30.6 & 90.2 & 72.0 & 56.5 & 94.2 & 39.4 & 73.9 & 55.6 & 65.38 \\
\rowcolor{aggrfill}\multicolumn{13}{l}{\quad\textit{Aggressive budget}, $\kbar\!\approx\!3$ ($\rbudget{=}0.38$)}\\
\rowcolor{aggrfill}\quad \fixedk{}, $k{=}3$ & 44.6 & 71.8 & 58.0 & 27.0 & 80.2 & 40.0 & 31.1 & 88.5 & 25.7 & 67.0 & 48.5 & 52.95 \\
\rowcolor{aggrfill}\quad \naee{}, $2.99/3.06$ & 41.3 & 68.4 & 59.7 & 27.6 & 75.6 & 23.2 & 16.9 & 81.1 & 13.7 & 57.4 & 39.4 & 45.85 \\
\rowcolor{aggrfill}\quad \diep{}, $3.36/3.07$$^{\dagger}$ & 38.9 & 67.6 & 54.0 & 23.8 & 58.4 & 3.3 & 6.0 & 86.4 & 4.6 & 59.5 & 39.9 & 40.22 \\
\rowcolor{aggrfill}\quad \diep{}-d, $3.01/3.04$ & 39.3 & 68.8 & 56.8 & 25.0 & 62.8 & 5.1 & 6.9 & 82.3 & 4.0 & 56.9 & 38.4 & 40.57 \\
\rowcolor{aggrfill}\quad \dynr{}, $2.97/3.08$ & 45.4 & 72.9 & 61.2 & 27.6 & 81.8 & 36.0 & 27.5 & 82.6 & 22.3 & 66.1 & 49.0 & 52.04 \\
\rowcolor{aggrfill}\quad \ban{}, $3.14/3.08$ & 46.1 & 72.8 & 61.5 & 25.8 & 88.0 & 42.9 & 34.0 & 90.4 & 28.6 & 68.8 & 50.5 & \textbf{55.40} \\
\bottomrule
\end{tabularx}
\end{table}

\begin{table}[t]
\centering
\scriptsize
\setlength{\tabcolsep}{0.92pt}
\caption{Every rule at both budget tiers on \textbf{Qwen3-Next-80B} ($\ktop{=}10$), per dataset. Uniform truncation is the first row of each tier so that the two can be read together. The budget beside a rule is its measured $\kbar$, generative\,/\,QA. A dagger marks a configuration that spends more than $10\%$ over its tier: it is shown for completeness but is not an iso-budget comparison and was not eligible for Table~\ref{tab:iso}. The best eligible mean is bold where it beats \fixedk{}, which is the row Table~\ref{tab:iso} carries.}
\label{tab:app:iso1}
\begin{tabularx}{\linewidth}{L cccc ccccc cc c}
\toprule
& \multicolumn{4}{c}{Knowledge QA}& \multicolumn{5}{c}{Math \& code reasoning}& \multicolumn{2}{c}{Knowledge \& general} & \\
\cmidrule(lr){2-5}\cmidrule(lr){6-10}\cmidrule(lr){11-12}
& ARC-c & ARC-e & WG & OBQA& MATH & AIME24 & AIME25 & GSM8K & LCB& MMLU-P & GPQA-D & Avg \\
\midrule
\quad unpruned, $k{=}10$ & 63.1 & 86.9 & 76.2 & 34.2 & 88.4 & 79.7 & 67.1 & 95.2 & 53.1 & 82.1 & 74.2 & 72.75 \\
\rowcolor{consfill}\multicolumn{13}{l}{\quad\textit{Conservative budget}, $\kbar\!\approx\!6$ ($\rbudget{=}0.60$)}\\
\rowcolor{consfill}\quad \fixedk{}, $k{=}6$ & 63.1 & 85.6 & 73.9 & 33.6 & 90.8 & 82.2 & 66.6 & 95.7 & 54.9 & 82.1 & 76.3 & 73.16 \\
\rowcolor{consfill}\quad \naee{}, $6.16/6.07$ & 59.8 & 83.6 & 72.0 & 32.6 & 89.6 & 80.1 & 63.0 & 95.1 & 49.1 & 81.2 & 69.2 & 70.48 \\
\rowcolor{consfill}\quad \diep{}, $6.24/6.13$ & 61.1 & 85.2 & 73.6 & 32.6 & 89.8 & 77.4 & 62.1 & 95.0 & 48.0 & 81.5 & 74.7 & 71.00 \\
\rowcolor{consfill}\quad \diep{}-d, $6.40/6.15$ & 60.3 & 84.9 & 73.4 & 31.8 & 90.4 & 79.1 & 64.0 & 95.0 & 53.1 & 81.3 & 72.7 & 71.45 \\
\rowcolor{consfill}\quad \dynr{}, $5.97/6.20$ & 62.2 & 85.7 & 73.0 & 35.6 & 89.6 & 83.4 & 67.9 & 95.1 & 50.9 & 81.9 & 77.3 & 72.96 \\
\rowcolor{consfill}\quad \ban{}, $6.18/6.08$ & 62.4 & 85.7 & 74.1 & 32.4 & 89.2 & 82.7 & 69.7 & 95.3 & 54.9 & 82.2 & 74.2 & 72.98 \\
\rowcolor{aggrfill}\multicolumn{13}{l}{\quad\textit{Aggressive budget}, $\kbar\!\approx\!4$ ($\rbudget{=}0.40$)}\\
\rowcolor{aggrfill}\quad \fixedk{}, $k{=}4$ & 55.9 & 82.2 & 67.2 & 30.2 & 88.0 & 74.8 & 58.8 & 93.1 & 49.1 & 80.7 & 72.7 & 68.43 \\
\rowcolor{aggrfill}\quad \naee{}, $4.17/4.05$ & 53.0 & 79.4 & 66.1 & 29.8 & 90.4 & 70.9 & 53.4 & 93.2 & 42.3 & 78.7 & 70.7 & 66.17 \\
\rowcolor{aggrfill}\quad \diep{}-d, $4.20/4.08$ & 52.4 & 80.2 & 66.8 & 31.0 & 92.4 & 71.7 & 53.0 & 94.1 & 42.9 & 80.1 & 67.2 & 66.53 \\
\rowcolor{aggrfill}\quad \dynr{}, $4.30/4.08$ & 57.8 & 82.3 & 70.8 & 33.0 & 89.2 & 78.5 & 62.2 & 95.0 & 54.3 & 81.1 & 69.7 & 70.35 \\
\rowcolor{aggrfill}\quad \ban{}, $4.27/4.18$ & 59.8 & 83.5 & 67.8 & 32.8 & 88.6 & 79.4 & 64.6 & 94.4 & 57.1 & 81.3 & 73.2 & \textbf{71.14} \\
\bottomrule
\end{tabularx}
\end{table}

\begin{table}[t]
\centering
\scriptsize
\setlength{\tabcolsep}{0.92pt}
\caption{Every rule at both budget tiers on \textbf{Ling-lite-1.5} ($\ktop{=}6$), per dataset. Uniform truncation is the first row of each tier so that the two can be read together. The budget beside a rule is its measured $\kbar$, generative\,/\,QA. A dagger marks a configuration that spends more than $10\%$ over its tier: it is shown for completeness but is not an iso-budget comparison and was not eligible for Table~\ref{tab:iso}. The best eligible mean is bold where it beats \fixedk{}, which is the row Table~\ref{tab:iso} carries.}
\label{tab:app:iso2}
\begin{tabularx}{\linewidth}{L cccc ccccc cc c}
\toprule
& \multicolumn{4}{c}{Knowledge QA}& \multicolumn{5}{c}{Math \& code reasoning}& \multicolumn{2}{c}{Knowledge \& general} & \\
\cmidrule(lr){2-5}\cmidrule(lr){6-10}\cmidrule(lr){11-12}
& ARC-c & ARC-e & WG & OBQA& MATH & AIME24 & AIME25 & GSM8K & LCB& MMLU-P & GPQA-D & Avg \\
\midrule
\quad unpruned, $k{=}6$ & 60.2 & 82.3 & 70.3 & 31.6 & 89.2 & 40.9 & 28.1 & 91.4 & 34.3 & 69.4 & 56.6 & 59.48 \\
\rowcolor{consfill}\multicolumn{13}{l}{\quad\textit{Conservative budget}, $\kbar\!\approx\!4$ ($\rbudget{=}0.67$)}\\
\rowcolor{consfill}\quad \fixedk{}, $k{=}4$ & 58.3 & 81.2 & 66.1 & 30.8 & 89.0 & 38.7 & 26.0 & 90.5 & 33.7 & 67.4 & 55.6 & 57.94 \\
\rowcolor{consfill}\quad \naee{}, $4.06/4.24$ & 56.3 & 81.5 & 65.7 & 29.4 & 87.4 & 33.2 & 23.3 & 87.9 & 30.3 & 61.9 & 53.5 & 55.49 \\
\rowcolor{consfill}\quad \diep{}, $4.09/4.08$ & 56.8 & 81.1 & 64.4 & 29.8 & 87.4 & 35.1 & 24.3 & 88.7 & 28.0 & 63.5 & 52.0 & 55.55 \\
\rowcolor{consfill}\quad \diep{}-d, $3.98/4.32$ & 57.4 & 81.6 & 67.8 & 30.4 & 86.2 & 34.1 & 25.8 & 87.9 & 32.0 & 62.8 & 58.1 & 56.74 \\
\rowcolor{consfill}\quad \dynr{}, $4.17/4.22$ & 58.4 & 82.0 & 68.4 & 30.4 & 90.2 & 40.1 & 26.8 & 90.5 & 32.0 & 67.3 & 58.6 & \textbf{58.61} \\
\rowcolor{consfill}\quad \ban{}, $3.97/4.07$ & 58.0 & 81.7 & 68.4 & 30.6 & 89.2 & 39.8 & 25.9 & 89.8 & 34.9 & 67.8 & 55.1 & 58.29 \\
\rowcolor{aggrfill}\multicolumn{13}{l}{\quad\textit{Aggressive budget}, $\kbar\!\approx\!2$ ($\rbudget{=}0.33$)}\\
\rowcolor{aggrfill}\quad \fixedk{}, $k{=}2$ & 38.9 & 69.6 & 57.4 & 23.0 & 70.4 & 5.0 & 8.4 & 72.4 & 8.0 & 42.3 & 38.4 & 39.44 \\
\rowcolor{aggrfill}\quad \naee{}, $2.43/2.28$$^{\dagger}$ & 46.9 & 73.7 & 57.2 & 26.4 & 74.2 & 12.0 & 11.9 & 78.8 & 18.9 & 48.5 & 46.5 & 45.00 \\
\rowcolor{aggrfill}\quad \diep{}, $2.69/2.25$$^{\dagger}$ & 43.9 & 73.1 & 57.7 & 25.8 & 77.0 & 15.1 & 15.9 & 84.1 & 17.1 & 51.1 & 40.9 & 45.61 \\
\rowcolor{aggrfill}\quad \diep{}-d, $2.08/2.05$ & 43.7 & 69.8 & 57.8 & 24.0 & 68.6 & 6.4 & 9.9 & 75.1 & 10.9 & 42.4 & 32.3 & 40.08 \\
\rowcolor{aggrfill}\quad \dynr{}, $2.04/2.10$ & 44.5 & 69.5 & 56.6 & 25.4 & 65.2 & 4.2 & 7.8 & 66.8 & 8.0 & 38.3 & 32.8 & 38.10 \\
\rowcolor{aggrfill}\quad \ban{}, $2.05$ & 42.2 & 71.2 & 57.1 & 24.6 & 70.4 & 10.2 & 11.5 & 75.1 & 14.9 & 47.5 & 41.9 & \textbf{42.42} \\
\bottomrule
\end{tabularx}
\end{table}

\begin{table}[t]
\centering
\scriptsize
\setlength{\tabcolsep}{0.92pt}
\caption{Every rule at both budget tiers on \textbf{GPT-OSS-20B} ($\ktop{=}4$), per dataset. Uniform truncation is the first row of each tier so that the two can be read together. The budget beside a rule is its measured $\kbar$, generative\,/\,QA. A dagger marks a configuration that spends more than $10\%$ over its tier: it is shown for completeness but is not an iso-budget comparison and was not eligible for Table~\ref{tab:iso}. The best eligible mean is bold where it beats \fixedk{}, which is the row Table~\ref{tab:iso} carries.}
\label{tab:app:iso3}
\begin{tabularx}{\linewidth}{L cccc ccccc cc c}
\toprule
& \multicolumn{4}{c}{Knowledge QA}& \multicolumn{5}{c}{Math \& code reasoning}& \multicolumn{2}{c}{Knowledge \& general} & \\
\cmidrule(lr){2-5}\cmidrule(lr){6-10}\cmidrule(lr){11-12}
& ARC-c & ARC-e & WG & OBQA& MATH & AIME24 & AIME25 & GSM8K & LCB& MMLU-P & GPQA-D & Avg \\
\midrule
\quad unpruned, $k{=}4$ & 45.3 & 77.5 & 66.9 & 27.2 & 90.6 & 73.7 & 71.7 & 86.2 & 58.3 & 74.3 & 64.6 & 66.94 \\
\rowcolor{consfill}\multicolumn{13}{l}{\quad\textit{Conservative budget}, $\kbar\!\approx\!3$ ($\rbudget{=}0.75$)}\\
\rowcolor{consfill}\quad \fixedk{}, $k{=}3$ & 44.8 & 76.7 & 65.4 & 27.4 & 91.0 & 77.4 & 74.5 & 87.0 & 61.7 & 73.6 & 65.7 & 67.75 \\
\rowcolor{consfill}\quad \naee{}, $2.97$ & 40.9 & 73.9 & 66.0 & 26.4 & 88.8 & 73.5 & 69.0 & 87.2 & 58.3 & 71.1 & 63.1 & 65.28 \\
\rowcolor{consfill}\quad \diep{}, $2.86$ & 39.2 & 73.0 & 63.9 & 22.4 & 85.2 & 50.4 & 44.2 & 86.2 & 32.0 & 67.7 & 57.1 & 56.47 \\
\rowcolor{consfill}\quad \diep{}-d, $2.99$ & 42.8 & 75.9 & 67.6 & 27.6 & 91.6 & 72.4 & 65.9 & 85.3 & 55.4 & 71.7 & 66.2 & 65.67 \\
\rowcolor{consfill}\quad \dynr{}, $2.97$ & 43.4 & 77.8 & 65.1 & 28.2 & 90.0 & 77.0 & 72.4 & 86.6 & 62.3 & 73.5 & 63.1 & 67.21 \\
\rowcolor{consfill}\quad \ban{}, $3.05$ & 44.8 & 76.8 & 66.2 & 28.0 & 92.4 & 73.1 & 70.4 & 86.9 & 60.0 & 73.4 & 60.6 & 66.59 \\
\rowcolor{aggrfill}\multicolumn{13}{l}{\quad\textit{Aggressive budget}, $\kbar\!\approx\!2$ ($\rbudget{=}0.50$)}\\
\rowcolor{aggrfill}\quad \fixedk{}, $k{=}2$ & 44.3 & 75.9 & 64.1 & 26.0 & 86.6 & 73.9 & 69.4 & 88.7 & 48.0 & 70.5 & 65.2 & 64.78 \\
\rowcolor{aggrfill}\quad \naee{}, $2.37$$^{\dagger}$ & 36.8 & 68.3 & 61.4 & 23.0 & 87.2 & 76.0 & 71.1 & 87.8 & 53.7 & 71.8 & 65.7 & 63.90 \\
\rowcolor{aggrfill}\quad \diep{}, $2.52$$^{\dagger}$ & 34.0 & 63.9 & 58.5 & 19.4 & 85.4 & 74.1 & 69.6 & 88.6 & 49.7 & 71.9 & 65.2 & 61.85 \\
\rowcolor{aggrfill}\quad \diep{}-d, $2.31$$^{\dagger}$ & 42.7 & 77.6 & 63.8 & 25.2 & 88.2 & 74.3 & 70.4 & 88.3 & 49.1 & 70.9 & 61.6 & 64.74 \\
\rowcolor{aggrfill}\quad \dynr{}, $2.07$ & 42.3 & 73.6 & 63.1 & 26.1 & 89.9 & 73.7 & 71.2 & 85.6 & 62.5 & 70.8 & 63.6 & \textbf{65.69} \\
\rowcolor{aggrfill}\quad \ban{}, $1.99$ & 44.8 & 76.2 & 65.0 & 26.2 & 86.2 & 71.5 & 66.0 & 87.2 & 40.6 & 68.2 & 63.1 & 63.18 \\
\bottomrule
\end{tabularx}
\end{table}

\section{Full results: the matched pairs}
\label{app:full:axes}

Table~\ref{tab:app:axes} gives the per-dataset \fixedk{} sweeps behind the two panels of
Figure~\ref{fig:axes}, and Tables~\ref{tab:app:modality:mm} and~\ref{tab:app:modality:qa} the two sides
of Figure~\ref{fig:modality}. The reference checkpoint appears in both so that each pair can be read on
its own.

The two suites of Figure~\ref{fig:modality} are matched in size rather than in content. The multimodal
side is MMMU \citep{mmmu}, MMBench \citep{mmbench}, MMStar \citep{mmstar}, ChartQA \citep{chartqa},
InfographicVQA \citep{infovqa}, TextVQA \citep{textvqa}, GQA \citep{gqa}, VizWiz \citep{vizwiz} and
MME-RealWorld \citep{mmerealworld}, scored with \citet{lmmseval}. The text side keeps the four
likelihood-scored datasets of \S\ref{sec:setup:data} and adds BoolQ \citep{boolq}, PIQA \citep{piqa},
Social IQa \citep{siqa}, MMLU \citep{mmlu} and HellaSwag \citep{hellaswag}, so that both models are read
on nine short-answer datasets apiece.

The figures average the four hardest generative datasets, which is a choice made for legibility; these
tables allow the same pairs to be read on any column. Two things visible here that the figures compress
are worth pointing at. The advantage that reasoning-mode post-training and parameter scale buy at three
experts is concentrated in the columns that need a long derivation, and is absent or slightly negative
on the four likelihood-scored ones, which is the split \S\ref{sec:axes} attributes it to. And the
multimodal collapse below four experts is not uniform across the multimodal suite either. ChartQA falls
furthest at every budget, and at two experts the three datasets that require transcribing text out of an
image are at or near zero, while the multiple-choice sets still return between a third and a half of
their unpruned scores. The pattern is the one \S\ref{sec:redundancy:regimes} describes, appearing inside
a single suite: what breaks first is the requirement to emit an exact string.

\begin{table}[t]
\centering
\scriptsize
\setlength{\tabcolsep}{0.92pt}
\caption{The two matched pairs of Figure~\ref{fig:axes} in full, per dataset. The reference checkpoint is repeated from Table~\ref{tab:app:fixedk} so that each pair can be read without turning back, and the budget tints are the same as in that table. Budgets absent from a block were not run: the thinking model has no $k{=}1$ and the $235$B model no $k{=}2$. Figure~\ref{fig:axes} averages the four hardest generative columns, AIME-24, AIME-25, LCB and GPQA-D.}
\label{tab:app:axes}
\begin{tabularx}{\linewidth}{L cccc ccccc cc c}
\toprule
& \multicolumn{4}{c}{Knowledge QA}& \multicolumn{5}{c}{Math \& code reasoning}& \multicolumn{2}{c}{Knowledge \& general} & \\
\cmidrule(lr){2-5}\cmidrule(lr){6-10}\cmidrule(lr){11-12}
& ARC-c & ARC-e & WG & OBQA& MATH & AIME24 & AIME25 & GSM8K & LCB& MMLU-P & GPQA-D & Avg \\
\midrule
\rowcolor{modelfill}\multicolumn{13}{l}{\textbf{Qwen3-30B-A3B-Instruct (reference)}, natively $\ktop{=}8$}\\
\rowcolor{consfill}\quad unpruned, $k{=}8$ & 60.8 & 85.0 & 73.2 & 32.0 & 89.6 & 74.1 & 61.4 & 94.4 & 41.1 & 74.2 & 56.6 & 67.49 \\
\rowcolor{consfill}\quad \fixedk{}, $k{=}7$ & 60.3 & 84.0 & 73.2 & 33.2 & 89.2 & 74.6 & 61.3 & 94.7 & 41.3 & 74.5 & 56.6 & 67.54 \\
\rowcolor{consfill}\quad \fixedk{}, $k{=}6$ & 59.2 & 82.8 & 71.3 & 31.4 & 90.4 & 72.8 & 59.3 & 95.0 & 44.6 & 74.4 & 53.5 & 66.79 \\
\rowcolor{consfill}\quad \fixedk{}, $k{=}5$ & 54.9 & 81.4 & 69.5 & 31.0 & 91.0 & 72.1 & 55.4 & 93.9 & 41.7 & 73.8 & 55.0 & 65.43 \\
\rowcolor{aggrfill}\quad \fixedk{}, $k{=}4$ & 51.3 & 77.2 & 65.1 & 31.2 & 87.8 & 62.5 & 47.9 & 93.3 & 40.6 & 72.3 & 54.0 & 62.11 \\
\rowcolor{aggrfill}\quad \fixedk{}, $k{=}3$ & 44.6 & 71.8 & 58.0 & 27.0 & 80.2 & 40.0 & 31.1 & 88.5 & 25.7 & 67.0 & 48.5 & 52.95 \\
\rowcolor{aggrfill}\quad \fixedk{}, $k{=}2$ & 28.7 & 51.0 & 50.6 & 21.4 & 24.6 & 0.0 & 0.0 & 22.9 & 1.1 & 22.1 & 27.3 & 22.70 \\
\rowcolor{aggrfill}\quad \fixedk{}, $k{=}1$ & 22.2 & 26.4 & 48.9 & 14.0 & 2.0 & 0.0 & 0.2 & 1.0 & 0.0 & 7.4 & 28.8 & 13.72 \\
\midrule
\rowcolor{modelfill}\multicolumn{13}{l}{\textbf{Qwen3-30B-A3B-Thinking}, natively $\ktop{=}8$}\\
\rowcolor{consfill}\quad unpruned, $k{=}8$ & 55.6 & 83.1 & 72.5 & 29.8 & 92.8 & 88.0 & 81.5 & 95.5 & 64.6 & 80.8 & 69.2 & 73.95 \\
\rowcolor{consfill}\quad \fixedk{}, $k{=}7$ & 55.2 & 82.2 & 72.1 & 30.0 & 93.2 & 88.1 & 80.4 & 95.7 & 62.9 & 80.9 & 70.2 & 73.72 \\
\rowcolor{consfill}\quad \fixedk{}, $k{=}6$ & 54.4 & 81.1 & 69.2 & 30.4 & 91.8 & 86.9 & 80.6 & 95.5 & 65.7 & 80.7 & 74.7 & 73.73 \\
\rowcolor{consfill}\quad \fixedk{}, $k{=}5$ & 51.2 & 78.3 & 67.9 & 29.4 & 93.4 & 84.7 & 78.5 & 95.8 & 59.4 & 80.4 & 70.2 & 71.75 \\
\rowcolor{aggrfill}\quad \fixedk{}, $k{=}4$ & 45.9 & 75.0 & 64.2 & 28.4 & 94.0 & 79.2 & 73.9 & 95.6 & 54.9 & 79.5 & 66.7 & 68.85 \\
\rowcolor{aggrfill}\quad \fixedk{}, $k{=}3$ & 40.1 & 67.8 & 56.3 & 28.0 & 90.0 & 61.8 & 50.1 & 93.3 & 30.3 & 75.1 & 62.1 & 59.54 \\
\rowcolor{aggrfill}\quad \fixedk{}, $k{=}2$ & 28.2 & 48.6 & 53.1 & 20.8 & 28.0 & 0.1 & 0.0 & 33.1 & 0.0 & 25.9 & 23.7 & 23.77 \\
\midrule
\rowcolor{modelfill}\multicolumn{13}{l}{\textbf{Qwen3-235B-A22B-Instruct}, natively $\ktop{=}8$}\\
\rowcolor{consfill}\quad unpruned, $k{=}8$ & 59.8 & 83.4 & 75.4 & 35.2 & 88.4 & 83.3 & 71.9 & 94.4 & 49.7 & 78.8 & 63.1 & 71.22 \\
\rowcolor{consfill}\quad \fixedk{}, $k{=}7$ & 59.6 & 82.6 & 74.7 & 33.2 & 89.2 & 80.0 & 69.0 & 94.9 & 49.7 & 78.4 & 63.6 & 70.45 \\
\rowcolor{consfill}\quad \fixedk{}, $k{=}6$ & 57.3 & 80.8 & 73.2 & 33.2 & 91.4 & 80.8 & 68.5 & 93.7 & 49.7 & 77.5 & 62.1 & 69.84 \\
\rowcolor{consfill}\quad \fixedk{}, $k{=}5$ & 56.7 & 80.3 & 70.6 & 32.4 & 93.2 & 78.3 & 66.9 & 93.4 & 52.0 & 77.5 & 60.1 & 69.22 \\
\rowcolor{aggrfill}\quad \fixedk{}, $k{=}4$ & 52.2 & 76.5 & 66.0 & 33.0 & 91.4 & 70.8 & 61.9 & 90.1 & 52.0 & 76.2 & 57.6 & 66.15 \\
\rowcolor{aggrfill}\quad \fixedk{}, $k{=}3$ & 42.1 & 68.3 & 58.0 & 29.6 & 86.4 & 56.9 & 48.8 & 89.2 & 38.3 & 72.6 & 61.6 & 59.25 \\
\rowcolor{aggrfill}\quad \fixedk{}, $k{=}1$ & 23.7 & 27.8 & 50.7 & 14.2 & 1.8 & 0.2 & 0.0 & 1.5 & 0.0 & 6.2 & 28.8 & 14.08 \\
\bottomrule
\end{tabularx}
\end{table}

\begin{table}[t]
\centering
\scriptsize
\setlength{\tabcolsep}{2pt}
\caption{Uniform truncation on the model behind the multimodal side of Figure~\ref{fig:modality}, per dataset. The two sides of that pair are nine datasets each and both are short answers, but they are not the same nine, so they are comparable only through the mean and only relative to each model's own unpruned row. Tints are the budget regions of Table~\ref{tab:app:fixedk}. Seven of the nine sets are the \texttt{lite} subsets.}
\label{tab:app:modality:mm}
\begin{tabularx}{\linewidth}{L ccccccccc c}
\toprule
\rowcolor{modelfill}\multicolumn{11}{l}{\textbf{Qwen3-VL-30B-A3B-Instruct, nine multimodal datasets, natively $\ktop{=}8$}}\\
\midrule
& MMMU & MMB & MMStar & ChartQA & InfoVQA & TextVQA & GQA & VizWiz & MME-RW & Avg \\
\midrule
\rowcolor{consfill}\quad unpruned, $k{=}8$ & 58.4 & 88.6 & 71.7 & 78.4 & 80.6 & 84.4 & 72.2 & 59.1 & 52.7 & 71.79 \\
\rowcolor{consfill}\quad \fixedk{}, $k{=}7$ & 58.2 & 90.2 & 71.1 & 76.4 & 79.1 & 84.8 & 71.4 & 58.7 & 54.2 & 71.57 \\
\rowcolor{consfill}\quad \fixedk{}, $k{=}6$ & 57.6 & 90.2 & 71.2 & 71.8 & 78.9 & 82.9 & 70.6 & 60.1 & 54.0 & 70.81 \\
\rowcolor{consfill}\quad \fixedk{}, $k{=}5$ & 56.0 & 88.6 & 69.6 & 68.4 & 78.5 & 82.8 & 70.6 & 58.8 & 52.9 & 69.58 \\
\rowcolor{aggrfill}\quad \fixedk{}, $k{=}4$ & 49.9 & 90.2 & 65.0 & 55.4 & 73.4 & 79.3 & 66.8 & 56.3 & 50.9 & 65.24 \\
\rowcolor{aggrfill}\quad \fixedk{}, $k{=}3$ & 47.3 & 83.3 & 55.3 & 28.6 & 56.0 & 58.7 & 54.6 & 44.3 & 39.9 & 52.00 \\
\rowcolor{aggrfill}\quad \fixedk{}, $k{=}2$ & 29.2 & 37.9 & 25.3 & 0.4 & 2.9 & 2.5 & 8.2 & 1.2 & 20.2 & 14.20 \\
\rowcolor{aggrfill}\quad \fixedk{}, $k{=}1$ & 26.2 & 15.9 & 23.3 & 0.0 & 0.0 & 0.0 & 0.0 & 0.0 & 19.7 & 9.46 \\
\bottomrule
\end{tabularx}
\end{table}

\begin{table}[t]
\centering
\scriptsize
\setlength{\tabcolsep}{2pt}
\caption{Uniform truncation on the model behind the text side of Figure~\ref{fig:modality}, per dataset. The two sides of that pair are nine datasets each and both are short answers, but they are not the same nine, so they are comparable only through the mean and only relative to each model's own unpruned row. Tints are the budget regions of Table~\ref{tab:app:fixedk}. The five datasets after OpenBookQA were measured for this comparison alone and are not part of the suite of \S\ref{sec:setup:data}.}
\label{tab:app:modality:qa}
\begin{tabularx}{\linewidth}{L ccccccccc c}
\toprule
\rowcolor{modelfill}\multicolumn{11}{l}{\textbf{Qwen3-30B-A3B-Instruct, nine likelihood-scored QA datasets, natively $\ktop{=}8$}}\\
\midrule
& ARC-c & ARC-e & WG & OBQA & BoolQ & PIQA & SIQA & MMLU & HSwag & Avg \\
\midrule
\rowcolor{consfill}\quad unpruned, $k{=}8$ & 60.8 & 85.0 & 73.2 & 32.0 & 88.7 & 80.1 & 54.1 & 80.1 & 61.2 & 68.36 \\
\rowcolor{consfill}\quad \fixedk{}, $k{=}7$ & 60.3 & 84.0 & 73.2 & 33.2 & 88.3 & 79.7 & 53.8 & 80.0 & 61.1 & 68.18 \\
\rowcolor{consfill}\quad \fixedk{}, $k{=}6$ & 59.2 & 82.8 & 71.3 & 31.4 & 86.7 & 79.6 & 53.5 & 79.2 & 60.9 & 67.18 \\
\rowcolor{consfill}\quad \fixedk{}, $k{=}5$ & 54.9 & 81.4 & 69.5 & 31.0 & 84.7 & 78.7 & 51.1 & 78.3 & 60.2 & 65.53 \\
\rowcolor{aggrfill}\quad \fixedk{}, $k{=}4$ & 51.3 & 77.2 & 65.1 & 31.2 & 82.1 & 76.6 & 47.6 & 75.3 & 58.4 & 62.76 \\
\rowcolor{aggrfill}\quad \fixedk{}, $k{=}3$ & 44.6 & 71.8 & 58.0 & 27.0 & 75.0 & 72.9 & 44.3 & 67.5 & 52.7 & 57.09 \\
\rowcolor{aggrfill}\quad \fixedk{}, $k{=}2$ & 28.7 & 51.0 & 50.6 & 21.4 & 53.8 & 62.0 & 36.6 & 34.4 & 39.9 & 42.04 \\
\rowcolor{aggrfill}\quad \fixedk{}, $k{=}1$ & 22.2 & 26.4 & 48.9 & 14.0 & 45.6 & 51.5 & 33.5 & 23.8 & 25.9 & 32.42 \\
\bottomrule
\end{tabularx}
\end{table}

\section{Throughput measurements}
\label{app:speed}

We time both backends through the generation call a user would make, not through a kernel benchmark.
Each measurement fixes a batch size, a prompt length and a number of generated tokens, and every
configuration measured at that shape sees the same random prompts. Generation runs to the requested
length rather than stopping early, and prefix caching is disabled in vLLM so that one measurement cannot
reuse work from another.

The two phases are timed separately by differencing rather than by instrumenting the backends. A run
with a single generated token gives prefill alone; the decode rate for a longer run at the same batch
size and prompt length is the remaining generated tokens divided by its wall time minus the matching
prefill time. Every run is executed once to warm up and then timed repeatedly at a fixed seed, and the
reported rate uses the median across trials.

\fixedk{} is applied here without the router patch of Appendix~\ref{app:setup}. We write a copy of the
model configuration with a smaller number of experts per token, so the serving stack selects fewer
experts natively and follows the same code path it would for a model published at that width. The
speedups therefore carry none of the overhead that a per-token allocation rule would add, which is the
distinction \S\ref{sec:speedup} draws between what the budget buys and what a rule would cost to
execute.

\section{Single-factor studies: further detail}
\label{app:axes}

\S\ref{sec:axes:modality} records routing concentration on prompt tokens, which raises the question of
whether the multimodal router's flatness belongs to the model or to the stage. The VL model's own
generated tokens answer it. Over its decode stage the natively selected eight experts hold $20.5\%$ of
the full softmax and the leading three hold $51.7\%$ of those eight, against $15.5\%$ and $49.8\%$ on
its prompts. Its routing is therefore somewhat more peaked when generating than when reading, but still
far flatter than the text model's $44.3\%$ and $59.8\%$. The difference between the two models is a
property of the models rather than of the stage at which we record it.

The size of that difference is also stable across depth. A $1000$-sample bootstrap over layers puts the
gap in top-eight mass at $-26.0$ points with a $95\%$ interval of $[-31.1,-21.2]$, and the router's
entropy at $+0.70$ nats higher in the multimodal model with an interval of $[+0.54,+0.85]$.

\end{document}